\documentclass[11pt]{article} 
\usepackage[margin=2cm,bottom=2.5cm]{geometry}
\usepackage[T1]{fontenc}
\usepackage{graphicx}
\usepackage{amsmath,amssymb,amsthm}
\usepackage{booktabs}
\usepackage{multirow}
\usepackage{xcolor}
\usepackage{colortbl}
\usepackage{hyperref}            
\usepackage{caption}             
\usepackage{algorithm}           
\usepackage{algpseudocode}       
\usepackage{float}               
\usepackage[section]{placeins}   
\usepackage[numbers,sort&compress]{natbib}
\usepackage{appendix}
\usepackage{fancyhdr}
\usepackage{authblk}  

\hypersetup{
  colorlinks=true,
  citecolor=blue,
  linkcolor=blue,
  urlcolor=blue
}

\algrenewcommand\algorithmicrequire{\textbf{Input:}}
\algrenewcommand\algorithmicensure{\textbf{Output:}}
\algrenewcommand\algorithmiccomment[1]{\hfill$\triangleright$ #1}

\title{\textbf{TopoFlow: Topography-aware Pollutant Flow Learning for High-Resolution Air Quality Prediction}}

\author[1,2]{Ammar Kheder\thanks{Corresponding author: \texttt{ammar.kheder@lut.fi}}}
\author[2,3]{Helmi Toropainen}
\author[2,3]{Wenqing Peng}
\author[4]{Samuel Antão}
\author[5]{Jia Chen}
\author[1,2,3]{Michael Boy}
\author[1,2]{Zhi-Song Liu}

\affil[1]{Department of Computational Engineering, Lappeenranta–Lahti University of Technology LUT, Finland}
\affil[2]{Atmospheric Modelling Centre Lahti, Lahti University Campus, Finland}
\affil[3]{Institute for Atmospheric and Earth System Research, University of Helsinki, P.O. Box 64, Helsinki 00014, Finland}
\affil[4]{Advanced Micro Devices (AMD), Munich, Germany}
\affil[5]{Technical University of Munich, Munich, Germany}

\date{}

\begin{document}
\let\cleardoublepage\relax

\maketitle
\thispagestyle{fancy}

\begin{abstract}
We propose TopoFlow (Topography-aware pollutant Flow learning), a physics-guided neural network for efficient, high-resolution air quality prediction. To explicitly embed physical processes into the learning framework, we identify two critical factors governing pollutant dynamics: topography and wind direction. Complex terrain can channel, block, and trap pollutants, while wind acts as a primary driver of their transport and dispersion. Building on these insights, TopoFlow leverages a vision transformer architecture with two novel mechanisms: topography-aware attention, which explicitly models terrain-induced flow patterns, and wind-guided patch reordering, which aligns spatial representations with prevailing wind directions. Trained on six years of high-resolution reanalysis data assimilating observations from over 1,400 surface monitoring stations across China, TopoFlow achieves a PM$_{2.5}$ RMSE of 9.71~$\mu$g/m$^3$, representing a 71--80\% improvement over operational forecasting systems and a 13\% improvement over state-of-the-art AI baselines. Forecast errors remain well below China's 24-hour air quality threshold of 75~$\mu$g/m$^3$ (GB 3095-2012), enabling reliable discrimination between clean and polluted conditions. These performance gains are consistent across all four major pollutants and forecast lead times from 12 to 96 hours, demonstrating that principled integration of physical knowledge into neural networks can fundamentally advance air quality prediction.
\end{abstract}

\textbf{Keywords:} Air pollution forecasting, Neural Networks, attention, air quality, China,
PM2.5

\section{Introduction}

Air pollution kills 8.1 million people annually, more than malaria, tuberculosis, and HIV/AIDS combined, yet forecasting systems still fail to predict, especially for pollutants that accumulate in complex terrain~\cite{hei2024}. Besides the driving force of meteorological parameters, e.g., wind directions, topography also plays a key role in shaping surface-level exposure: mountains impede horizontal dispersion, valleys channel airflow along preferred directions, and enclosed basins trap pollutants under stable atmospheric conditions. These terrain effects have contributed to severe historical pollution events, from the 1930 Meuse Valley disaster~\cite{nemery2001} to persistent wintertime haze episodes in China's Sichuan Basin and Guanzhong Plain~\cite{bei2016, wang2018}. The governing physics is well established, as terrain modulates both pollutant dispersion and the near-surface wind field that drives advection~\cite{whiteman2000}, yet data-driven forecasting approaches typically lack explicit terrain awareness, limiting their ability to capture these effects.

Current numerical chemical transport models struggle to resolve terrain-pollution interactions. Global systems, such as CAMS, operate at approximately 0.75° resolution, insufficient to capture valley-scale circulations and basin confinement occurring at 10 to 50 km scales~\cite{ge2020}. Regional systems like CUACE achieve PM$_{2.5}$ errors of 34 to 48~$\mu$g/m$^3$ at 24-hour lead time, representing 45--64\% of China's regulatory threshold (75~$\mu$g/m$^3$, GB~3095-2012)~\cite{GB3095_2012}, due to coarse grids, uncertain emissions, and computational constraints~\cite{dai2019,qi2022}, with systematic errors persisting over mountainous western China and the Tibetan Plateau~\cite{ali2022}.
Even WRF-Chem, the most widely used regional model in China, 
shows that basin topography alone can enhance PM$_{2.5}$ concentrations by up to 
12\%~\cite{jiang2020terrain}, yet exhibits substantial forecast biases that 
require combined data assimilation and deep-learning correction to reduce RMSE 
by 28--62\%~\cite{peng2024wrfchem}.
Unlike benchmark-driven computer vision tasks where marginal gains have limited practical significance, air quality forecasting is a domain where each improvement directly translates to public health outcomes: earlier advisories, better hospital preparedness, and lives saved.

Recent AI weather forecasting models~\cite{climax, graphcast, pangu, aurora} rival numerical prediction at orders of magnitude lower computational cost, achieving comparable results. However, only a few works have studied the air pollution predictions thoroughly. Furthermore, they have not explicitly optimized the neural network architectures to learn physical constraints. For instance, built on the vision transformer~\cite{vit}, ClimaX relies on generic attention mechanisms that learns the spatial correlations from the large-scale training data. Some approaches model station measurements as graphs and approximate the air pollution prediction as ODE systems without considering physical constraints \cite{airphynet, xu2023, Kheder2025}. No existing method embeds the physical relationship between terrain geometry and atmospheric transport directly into the model architecture.

In this work, we are interested in encoding physical constraints into the neural networks explicitly for accurate air pollution prediction. Particularly, we focus on two physical factors: 1) terrain imposes directionally asymmetric transport where downslope dispersion is favored over upslope advection, and 2), pollutants follow wind streamlines rather than arbitrary spatial patterns. Our keen focus is high-resolution air pollution prediction (15~km~$\times$~15~km), hence we choose vision transformer~\cite{vit} among many network architectures. The reason is that it can encode the 2D map as localized patch tokens for global feature extraction via the attention~\cite{attention}. Based on that, we introduce two novel mechanisms: wind-guided patch reordering, which aligns grid cells (patch tokens) with dominant advective directions, and topography-aware attention, which encodes altitude differences as flow barriers within the attention computation. We denote it as TopoFlow, topography-aware pollutant Flow learning. Using six years of high-resolution air quality reanalysis dataset over China~\cite{kong2021}, we show that TopoFlow achieves the RMSE of 9.71 $\mu$g/m$^3$ on PM$_{2.5}$ prediction, representing a 71 to 80\% improvement over operational systems (CAMS, CUACE) and 13\% improvement over state-of-the-art AI baselines including ClimaX~\cite{climax} and AirCast~\cite{aircast}. During the severe November 2018 Beijing haze episode, TopoFlow reduces error by 92\% compared to CAMS, with transport dynamics independently validated by Lagrangian trajectory analysis~\cite{foreback2025}. These results demonstrate that embedding physical constraints into the network architecture design enables accurate prediction of geography-atmosphere interactions that both numerical models and existing deep learning approaches fail to capture.

\section{Results}

\subsection*{Model design and experimental setting}

\begin{figure*}[t]
    \centering
    \includegraphics[width=\textwidth]{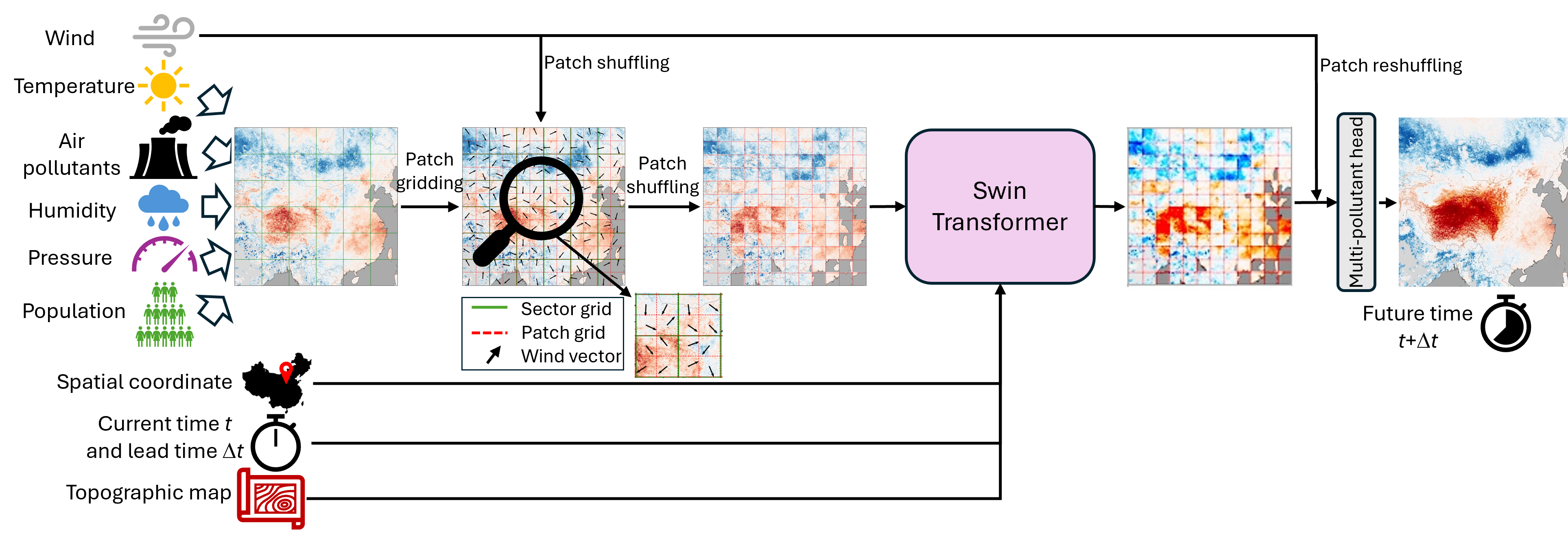}
    \caption{\textbf{TopoFlow architecture for physics-guided air quality prediction.} The model takes as input concentrations of six air pollutants, major meteorological data, population density, spatial coordinates, time stamps, and a topographic map, and outputs pollutant concentrations at lead times from 12 to 96 hours. All input data are stacked into a multi-layer 2D map, then cropped into non-overlapping patches. TopoFlow shuffles patch order based on the wind field within each sector, then processes patches through a Swin Transformer backbone. The topographic map introduces attention bias for topography-aware feature representation.}
    \label{fig:network}
\end{figure*}

The TopoFlow is proposed as a physics-guided Vision Transformer for multi-pollutant air quality forecasting. The model significantly improves the Transformer architecture~\cite{climax,vit,swin} to learn spatiotemporal feature correlations for pollution forecasting (Fig.~\ref{fig:network}).

Originally, the vision transformer takes the input 2D map, which is cropped into patches and sorted in raster order, then processed via the attention mechanism for global feature extraction. Our key innovation contains two parts: (1) \textit{wind-guided patch reordering} to align patch sequences with the wind field, whose physical motivation from advection-dominated transport theory is detailed in Appendix~A, and (2) \textit{topography-aware attention bias} to incorporate terrain-induced flow barriers.

For wind-guided patch reordering, we crop the input data into regional sectors. Within each sector, we further crop into smaller patches. The patch order within each sector is determined by the wind direction (see Algorithm~B1 in Appendix~B), and the attention block aligns the patches with the winds for physics-biased learning. For topography-aware attention bias, the topographic map is used as a physical bias to weight different regions based on elevation. This bias is inserted into the attention calculation to penalize high-altitude regions with less influence on pollution transport.

TopoFlow operates on a $128 \times 256$ spatial grid at $0.25^\circ$ resolution covering China, forecasting six pollutants (PM$_{2.5}$, PM$_{10}$, SO$_2$, NO$_2$, CO, O$_3$) with resolution 15~km~$\times$~15~km at four horizons (+12h, +24h, +48h, +96h). The model input is a multivariate spatiotemporal field $\mathbf{X} \in \mathbb{R}^{H \times W \times V_{\text{in}}}$ with $H=128$, $W=256$, and $V_\text{in}=15$, comprising: meteorological fields (5 variables: horizontal wind $u$ and $v$, temperature, relative humidity, surface pressure), pollutant concentrations (6 species, from CAQRA reanalysis~\cite{kong2021}), spatial coordinates (latitude, longitude), and static geographic features (topography, population density); a complete specification of all input variables is provided in Supplementary Table~1.

We use the reanalysis data from the Chinese Air Quality Reanalysis (CAQRA)~\cite{kong2021} covering China from 2013 to 2018. Topographic data is obtained from ETOPO1~\cite{amante2009}. Besides these, we also add the mean population density from the Gridded Population of the World dataset~\cite{ciesin2018}. All meteorological and pollutant fields are standardized using $z$-score normalization \cite{bishop2006} based on training statistics, while elevation and population are normalized to range [0, 1] (Supplementary Section~1.1 details all preprocessing and normalization steps). The dataset is split temporally: 2013 to 2016 for training, 2017 for validation, and 2018 for testing (Supplementary Table~4 details the test set sample distribution across seasons, times of day, and days of the month). To evaluate generalization to independent observations, we additionally validate against 637 surface monitoring stations from the OpenAQ network \cite{openaq2024} across China for 2019, ensuring complete separation between training data and evaluation.

The model contains 52.5 million learnable parameters (see Supplementary Table~3 for the complete hyperparameter configuration). We train TopoFlow for 60 epochs (20,000 optimization steps) on 128 AMD MI250X GPUs hosted on the LUMI supercomputer\footnote{\url{https://lumi-supercomputer.eu/}} with a batch size of 512, completing in approximately 8 days (Supplementary Section~1.2 provides full computational infrastructure details).

\subsection*{Multi-pollutant forecasting performance}

\begin{figure}[!htb]
    \centering
    \includegraphics[width=\textwidth]{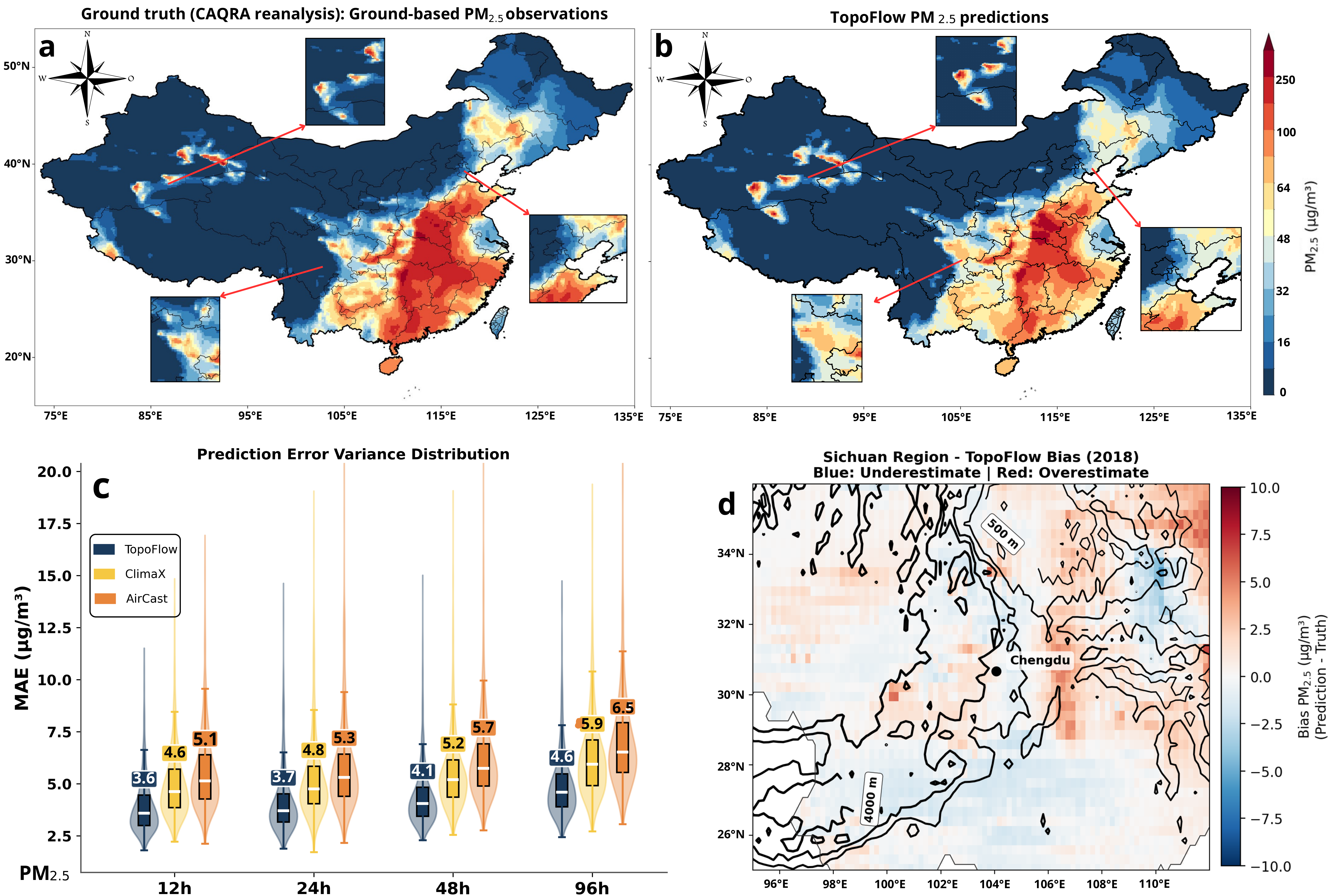}
    \caption{\textbf{Overall performance of air pollution 
    prediction.} \textbf{(a)}, Ground truth (CAQRA reanalysis) 
    PM$_{2.5}$ observations. \textbf{(b)}, TopoFlow PM$_{2.5}$ 
    predictions. \textbf{(c)}, Prediction error 
    ($|\hat{y} - y|$, where $\hat{y}$ is the model prediction and 
    $y$ the CAQRA reanalysis) distribution across lead times 
    comparing TopoFlow, ClimaX, and AirCast. Box plots indicate 
    median (middle line), 25th and 75th percentile (box), and 5th 
    and 95th percentile (whiskers). \textbf{(d)}, Spatial 
    distribution of TopoFlow bias in Sichuan (with complex 
    terrains), showing underestimation (blue) in the basin interior 
    and overestimation (red) near elevated margins, consistent with 
    residual difficulty in resolving sharp terrain-induced 
    concentration gradients at the plateau-basin interface.}
    \label{fig:teaser}
\end{figure}

In this section, we evaluate the performance of TopoFlow by comparing it against state-of-the-art machine learning models and operational forecasting systems (Fig.~\ref{fig:teaser}).

TopoFlow was trained on CAQRA, a 6-year high-resolution (15~km) 
gridded dataset that assimilates surface observations from over 
1,400 monitoring stations operated by the China National 
Environmental Monitoring Center (CNEMC). To ensure a fair 
comparison, ClimaX and AirCast were trained on the same 
CAQRA training data (2013--2016) as TopoFlow, while Aurora was 
pre-trained on ERA5 and CAMS reanalysis~\cite{aurora} 
(Supplementary Table~2). TopoFlow achieves PM$_{2.5}$ RMSE of 
9.71~$\mu$g/m$^3$, representing 13\% improvement over ClimaX 
(11.16~$\mu$g/m$^3$) and 21\% improvement over AirCast 
(12.25~$\mu$g/m$^3$). Similar performance gains are observed 
for PM$_{10}$, where TopoFlow (17.25~$\mu$g/m$^3$) outperforms 
all baselines. For nitrogen dioxide (NO$_2$) and sulfur dioxide 
(SO$_2$), TopoFlow demonstrates particularly strong performance 
at shorter lead times, with RMSE values of 
7.79~$\mu$g/m$^3$ and 1.44~$\mu$g/m$^3$ at 12-hour forecasts, 
respectively. A comprehensive breakdown across all pollutants 
and lead times is provided in Supplementary Tables~5--6.

\subsection*{Validation against independent ground measurements}

\begin{figure}[!htb]
    \centering
    \includegraphics[width=\textwidth]{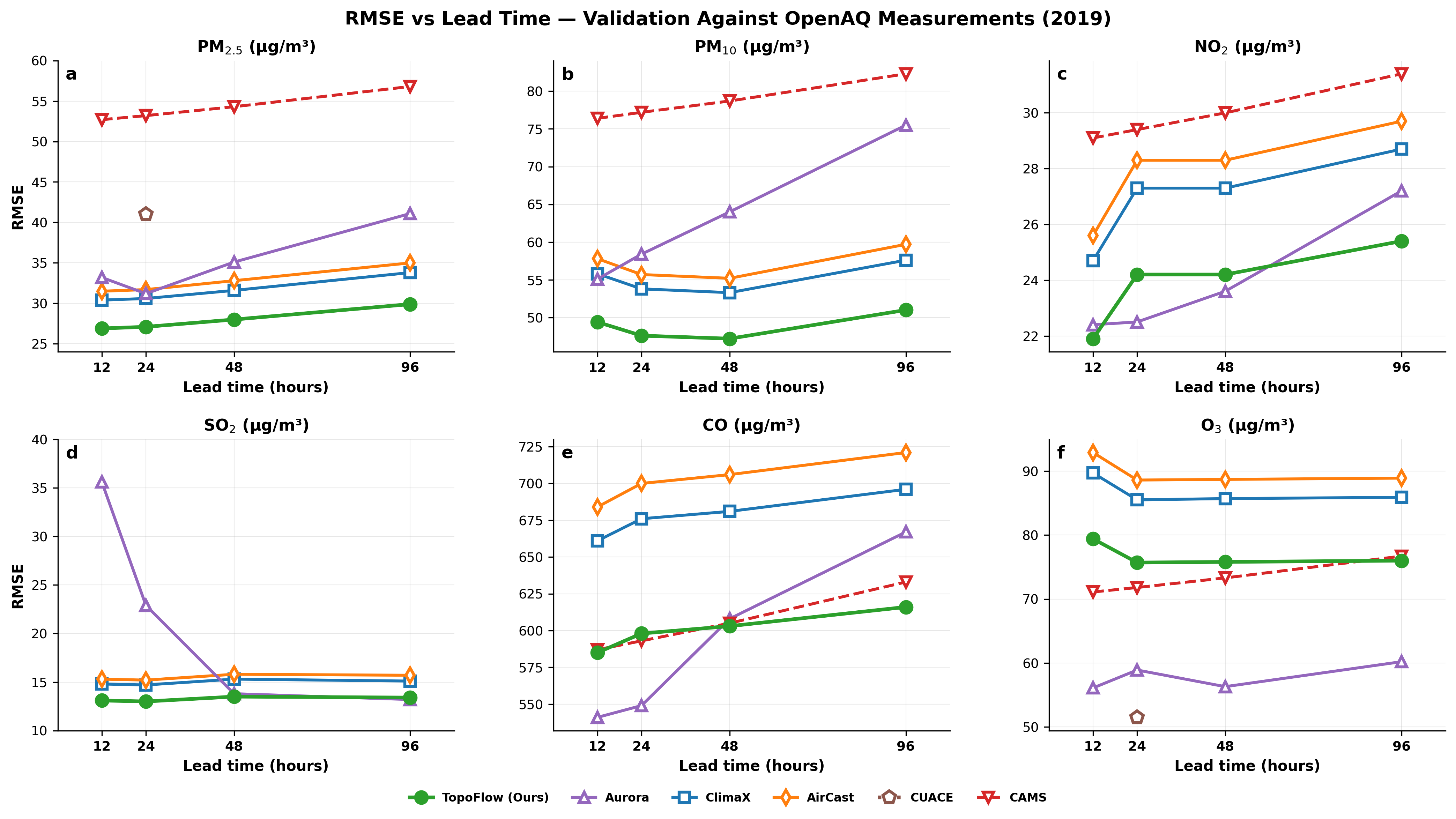}
    \caption{\textbf{Forecast skill as a function of lead time for six air pollutants.} RMSE validated against OpenAQ stations across China for 2019. \textbf{a}, PM$_{2.5}$. \textbf{b}, PM$_{10}$. \textbf{c}, NO$_2$. \textbf{d}, SO$_2$. \textbf{e}, CO. \textbf{f}, O$_3$. TopoFlow (green) achieves the lowest errors for particulate matter and  NO$_2$. Aurora (purple) shows superior performance for O$_3$ and CO, which require three-dimensional atmospheric representation to capture stratospheric intrusions and vertical transport.}
    \label{fig:rmse_leadtime}
\end{figure}

We validated TopoFlow forecasts against 637 independent surface monitoring stations from the OpenAQ network across China for 2019 (Fig.~\ref{fig:rmse_leadtime}).

This strategy ensures complete separation between training data (CAQRA reanalysis) and evaluation data (direct observations), providing an unbiased assessment of real-world forecast skill. We compared TopoFlow against five established systems: 
Aurora~\cite{aurora}, the state-of-the-art AI foundation model 
for Earth system prediction; ClimaX~\cite{climax}, a climate 
foundation model; AirCast~\cite{aircast}, a deep learning 
approach for air quality; and two operational numerical 
chemistry systems, CAMS from ECMWF~\cite{inness2019cams} and 
CUACE from the China Meteorological 
Administration~\cite{dai2019,qi2022}. A detailed comparison of 
model architectures, training data, and physics representations 
is provided in Supplementary Table~2.

TopoFlow achieves the lowest errors for particulate matter 
(PM$_{2.5}$, PM$_{10}$), nitrogen dioxide (NO$_2$), and sulfur 
dioxide (SO$_2$) across all forecast horizons from 12 to 96 
hours. For PM$_{2.5}$, TopoFlow maintains RMSE values between 
27 and 30~$\mu$g/m$^3$, representing 22\% improvement over 
ClimaX and 45\% improvement over CAMS. Across all species, 
TopoFlow errors remain below 20\% of the corresponding 
GB~3095-2012 regulatory thresholds~\cite{GB3095_2012} (Supplementary Table~7), whereas 
CAMS errors for SO$_2$ and PM$_{10}$ approach or exceed half 
of their respective limits (Supplementary Table~7). Aurora 
demonstrates superior skill for ozone (O$_3$) and CO, achieving RMSE values 
approximately 25\% lower than TopoFlow. This performance gap reflects 
architectural differences as well as the longer atmospheric lifetimes of
ozone and CO compared to PM, NO$_2$, and SO$_2$: longer-lived species are 
governed by large-scale transport that Aurora captures through its 
three-dimensional multi-pressure-level representation, whereas short-lived 
pollutants are dominated by local emissions and terrain-driven dispersion 
where TopoFlow's surface-level physics guidance excels.

Volatile organic compounds (VOCs) and nitrogen oxides (NO$_\text{x}$) are central to tropospheric ozone chemistry. Surface ozone concentrations depend on photochemical production through NO$_\text{x}$-VOC interactions, but are also influenced by stratospheric intrusions where ozone-rich stratospheric air descends through tropopause folding to the surface~\cite{lin2015stratospheric,knowland2017stratospheric}. These intrusions contribute 5 to 15\% of tropospheric ozone and are pronounced over mountainous terrain~\cite{zhao2021stratospheric,lefohn2012stratospheric}. Similarly, CO undergoes vertical redistribution through deep convection during monsoon seasons~\cite{kar2004vertical,park2009transport}. Aurora's 3D Swin Transformer architecture captures these vertical processes explicitly~\cite{aurora}. TopoFlow's superior performance on particulate matter despite using only surface data is physically interpretable: PM$_{2.5}$ and PM$_{10}$ concentrations are predominantly determined by surface emissions, boundary layer dynamics, and horizontal advection~\cite{advection,whiteman2000}.

\subsection*{Seasonal and spatial validation}

\begin{figure}[!htb]
    \centering
    \includegraphics[width=\textwidth]{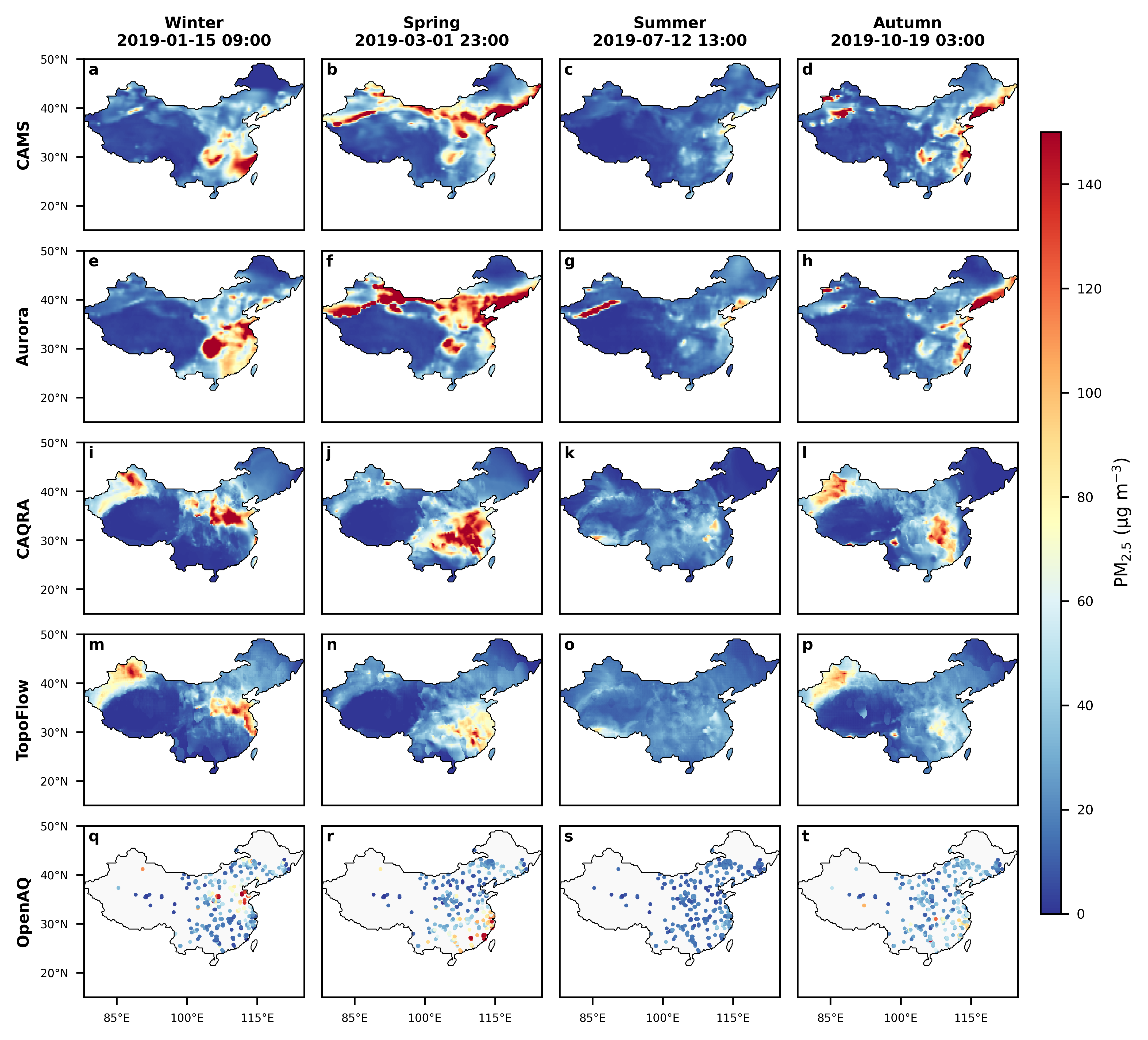}
    \caption{\textbf{Seasonal PM$_{2.5}$ distribution from forecasts, reanalysis, and observations.} \textbf{(a--d)}, CAMS forecasts. \textbf{(e--h)}, Aurora predictions. \textbf{(i--l)}, CAQRA reanalysis. \textbf{(m--p)}, TopoFlow predictions. \textbf{(q--t)}, OpenAQ measurements. Columns: Winter (15 January 2019), Spring (1 March 2019), Summer (12 July 2019), Autumn (19 October 2019). TopoFlow achieves lowest RMSE (25.9 $\mu$g/m$^3$) against independent stations, outperforming CAQRA (37.0 $\mu$g/m$^3$), CAMS (44.1 $\mu$g/m$^3$), and Aurora (49.8 $\mu$g/m$^3$).  Relative to China's 24-hour 
PM$_{2.5}$ threshold of 75~$\mu$g/m$^3$ 
(GB~3095-2012)~\cite{GB3095_2012}, only TopoFlow and CAQRA 
maintain errors below 50\% of the regulatory limit.}
    \label{fig:spatial_validation}
\end{figure}

Fig.~\ref{fig:spatial_validation} compares spatial PM$_{2.5}$ predictions from four systems against ground observations across seasonal conditions.

CAMS forecasts and Aurora predictions exhibit concentration maxima exceeding 120~$\mu$g/m$^3$ along the Himalayan foothills and North China Plain during winter and spring (\textbf{(h)} and \textbf{(l)} in Fig.~\ref{fig:spatial_validation}), systematically overestimating surface concentrations consistent with previous evaluations~\cite{wu2020cams,ali2022}. 
TopoFlow predictions exhibit smoother spatial fields, a characteristic shared by other AI weather models, including GraphCast~\cite{graphcast} and Pangu-Weather~\cite{pangu}, that arise from neural network regression under mean squared error optimization.

Quantitative validation confirms that TopoFlow achieves a mean RMSE of 25.9~$\mu$g/m$^3$ against OpenAQ stations, outperforming CAQRA (37.0~$\mu$g/m$^3$), CAMS (44.1~$\mu$g/m$^3$), and Aurora (49.8~$\mu$g/m$^3$). Seasonal analysis reveals consistent advantages: winter conditions yield comparable errors between TopoFlow (38.5~$\mu$g/m$^3$) and CAQRA (38.9~$\mu$g/m$^3$), while CAMS (51.8~$\mu$g/m$^3$) and Aurora (59.3~$\mu$g/m$^3$) overestimate substantially. Spring presents challenging conditions with dust transport and biomass burning; TopoFlow (34.2~$\mu$g/m$^3$) reduces RMSE by 38\% relative to CAQRA (55.4~$\mu$g/m$^3$). Summer monsoon conditions produce the lowest errors across all systems due to strong dispersion, with TopoFlow reaching 10.5~$\mu$g/m$^3$.

\begin{figure}[!htb]
    \centering
    \includegraphics[width=\textwidth]{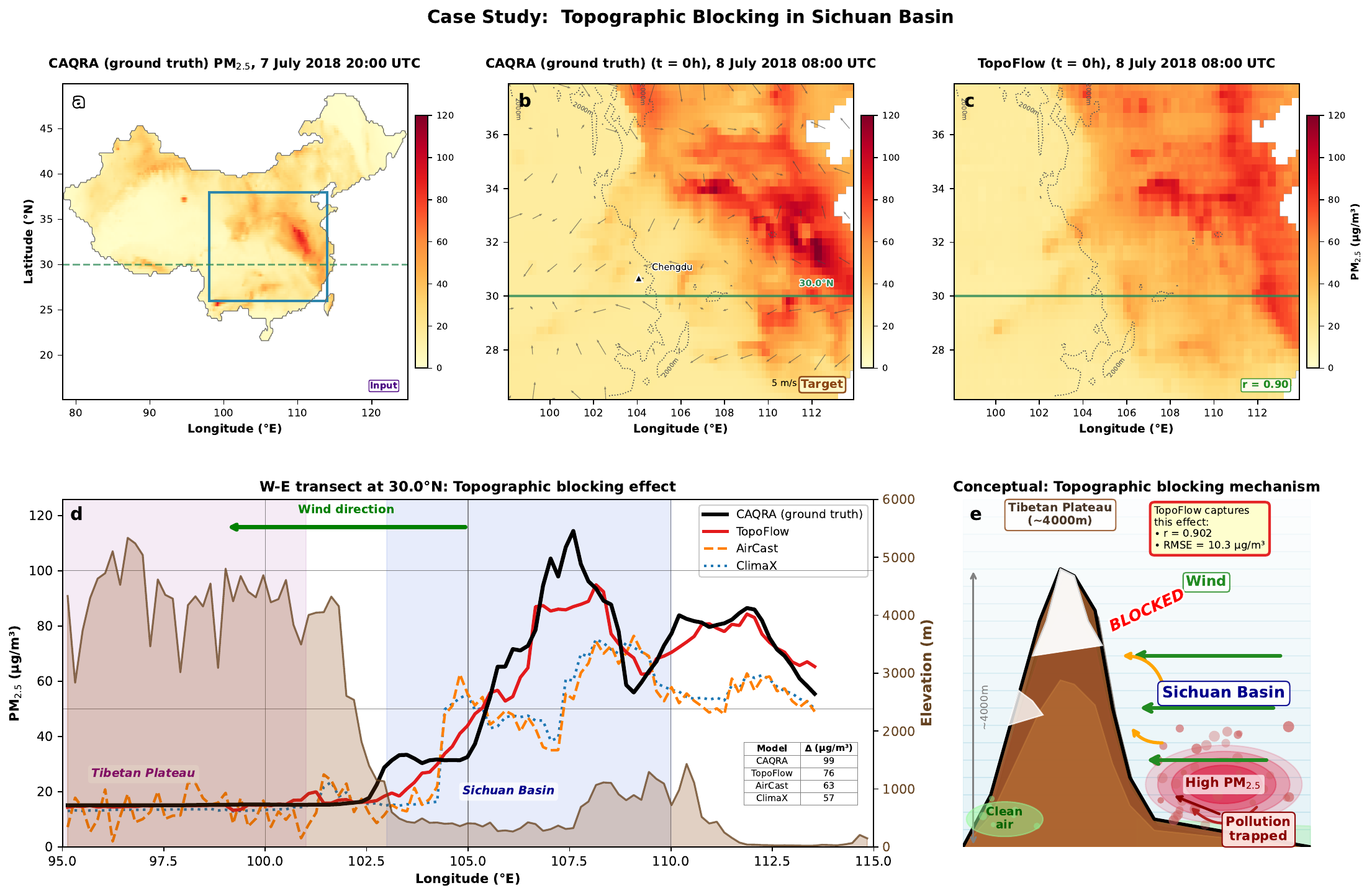}
    \caption{\textbf{Topographic blocking in Sichuan Basin.} 
    \textbf{(a)}, CAQRA (ground truth) PM$_{2.5}$ distribution 
    across China at 7 July 2018, 20:00 UTC, with the study 
    region marked by the green box. \textbf{(b)}, CAQRA (ground 
    truth) PM$_{2.5}$ concentrations and wind vectors at forecast 
    time (8 July 2018, 08:00 UTC) within the Sichuan Basin. The 
    green line indicates 30.0°N transect. \textbf{(c)}, TopoFlow 
    12-hour prediction achieving spatial correlation r = 0.90. 
    \textbf{(d)}, West-east transect along 30.0°N comparing CAQRA 
    (black), TopoFlow (red), AirCast (orange dashed), and ClimaX 
    (blue dotted) against elevation profile (gray shading). Inset 
    table reports the terrain-induced concentration gradient 
    $\Delta = C_{\text{basin,max}} - C_{\text{plateau,min}}$ for 
    each model. \textbf{(e)}, Schematic of topographic blocking 
    mechanism: the Tibetan Plateau blocks westerly winds, trapping 
    pollutants within the basin.}
    \label{fig:figure_study}
\end{figure}

\subsection*{Topographic blocking in Sichuan Basin}

To demonstrate that TopoFlow's topography-aware attention mechanism 
captures the fundamental physics of terrain-atmosphere interactions, 
we examine a summer pollution episode in the Sichuan Basin 
(Fig.~\ref{fig:figure_study}).

The Sichuan Basin, a deep sedimentary depression in southwestern 
China enclosed by the Tibetan Plateau ($\sim$4,000~m) to the west, 
functions as a natural trap for atmospheric pollutants where 
terrain-induced blocking governs pollution accumulation patterns 
that conventional deep learning approaches systematically fail to 
reproduce.

The 12-hour forecast initialized on 7 July 2018 reveals the 
characteristic signature of topographic blocking: concentrations 
exceeding 100~$\mu$g/m$^3$ accumulate within the basin interior 
while clean air (<20~$\mu$g/m$^3$) persists over the elevated 
western margins. Wind vectors indicate westerly flow at 
approximately 5~m/s, confirming active transport conditions where 
clean air masses from the Tibetan Plateau encounter the basin's 
western rim. The west-east transect along 30.0°N provides 
quantitative validation: the CAQRA reanalysis PM$_{2.5}$ profile 
exhibits a sharp concentration gradient 
$\Delta = C_{\text{basin,max}} - C_{\text{plateau,min}}$ = 
99~$\mu$g/m$^3$ at the elevation transition from the Tibetan 
Plateau to the basin floor. TopoFlow captures 77\% of this 
gradient ($\Delta$ = 76~$\mu$g/m$^3$, r~=~0.902, 
RMSE~=~10.3~$\mu$g/m$^3$), while AirCast 
($\Delta$ = 63~$\mu$g/m$^3$, 64\%) and ClimaX 
($\Delta$ = 57~$\mu$g/m$^3$, 58\%) substantially underestimate 
the terrain-induced concentration discontinuity despite 
demonstrated skill in flat terrain. This topographic blocking 
extends to all six pollutants 
(Fig.~A1): TopoFlow captures 
concentration gradients for PM$_{10}$ ($R = 0.91$), NO$_2$ 
($R = 0.84$), CO ($R = 0.84$), and SO$_2$ ($R = 0.73$), while 
ozone exhibits an inverse gradient consistent with nitric oxide 
(NO) titration within the polluted basin ($R = 0.27$).

\subsection*{November 2018 Beijing severe haze episode}

\begin{figure*}[t!]
\centering
\includegraphics[width=0.95\textwidth]{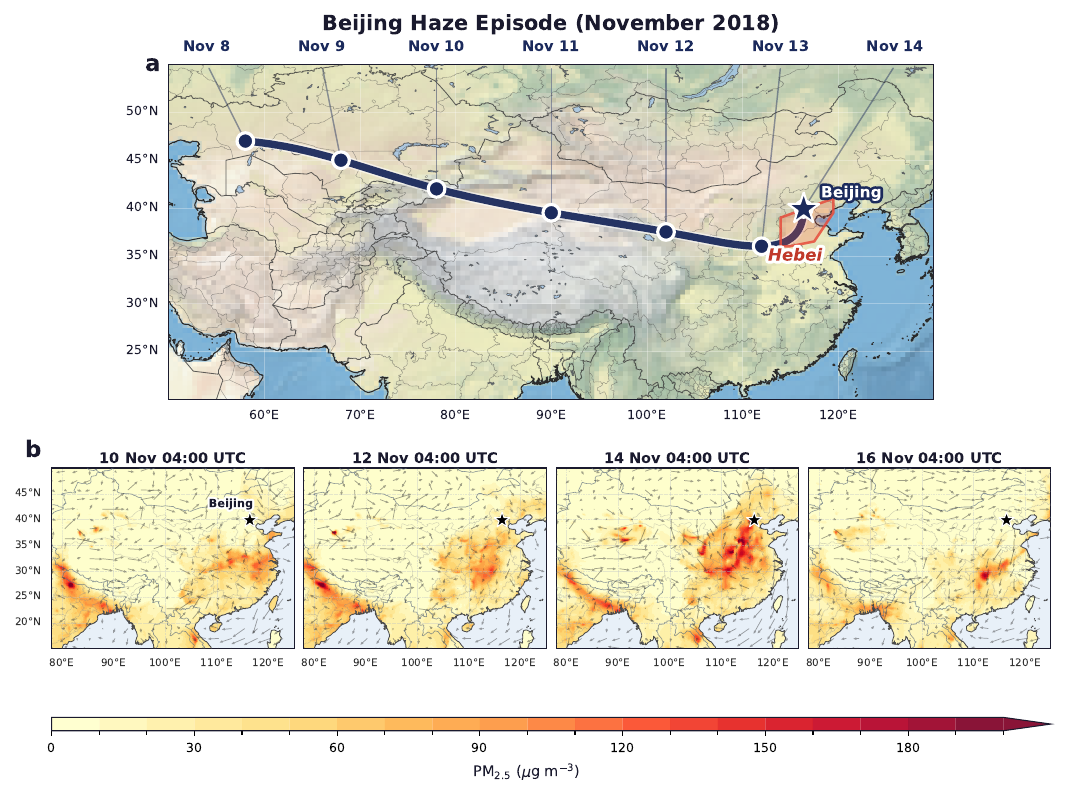}
\caption{\textbf{November 2018 Beijing haze episode.}
\textbf{(a)}, Seven-day back-trajectory arriving at Beijing on 14 November 2018, adapted from FLEXPART-SOSAA analysis in Foreback et al.~\cite{foreback2025}. The trajectory originates from Central Asia and passes through the Xinjiang region before traversing Hebei province and arriving at Beijing. Dates along the trajectory indicate the air mass position at daily intervals from 8 to 14 November.
\textbf{(b)}, TopoFlow-predicted PM$_{2.5}$ concentration fields on 10, 12, 14, and 16 November 2018 with ERA5 wind vectors overlaid. Beijing is marked with a star. The sequence shows the progressive buildup of pollution across northern China during the episode (10--14 November) and the rapid clearing on 16 November following the passage of a clean northwesterly air mass.}
\label{fig:case_study}
\end{figure*}

To evaluate TopoFlow's capacity to capture physical dynamics during extreme pollution events, we analyze the severe haze episode affecting Beijing from 11 to 15 November 2018 (Fig.~\ref{fig:case_study}).

This episode was characterized by PM$_{2.5}$ concentrations exceeding 200~$\mu$g/m$^3$ at monitoring stations in Beijing, more than 13 times the WHO 24-hour guideline value of 15~$\mu$g/m$^3$~\cite{foreback2025}. The episode was driven by a persistent blocking high-pressure system over the Beijing region, resulting in stagnant conditions with light southerly winds that allowed pollutant concentrations to accumulate over several days~\cite{foreback2025}. Using the FLEXPART-SOSAA Lagrangian modeling system, Foreback et al.~\cite{foreback2025} demonstrated that approximately 77\% of particles observed in Beijing during this episode originated from outside the city, with back-trajectories indicating that up to 5\% of primary particle mass and 7\% of secondary organic aerosol mass were transported from the Xinjiang region in western China, roughly 2000~km to the west. The trajectory arriving on 14 November (Fig.~\ref{fig:case_study}\textbf{(a)}) was classified as haze upon reaching central China, approximately 
three days before arriving at Beijing, and as severe haze more than one day before arrival upon entering Hebei province~\cite{foreback2025}.

The TopoFlow predictions 
(Fig.~\ref{fig:case_study}\textbf{(b)}) capture the key 
spatiotemporal features of this episode independently identified 
by Foreback et al.~\cite{foreback2025}: the progressive 
accumulation of PM$_{2.5}$ across the North China Plain during 
10--14 November, the peak concentrations on 14 November 
extending from Hebei province into Beijing, and the abrupt 
clearing on 16 November when a synoptic weather shift brought 
clean air from the northwest. The wind vectors in 
panel~\textbf{(b)} show the transition from weak, variable winds 
during the episode to stronger northwesterly flow during the 
clearing phase, consistent with the meteorological analysis 
in~\cite{foreback2025}. Over northern China on 14 November, 
TopoFlow achieves RMSE of 5.88~$\mu$g/m$^3$ and MAE of 
3.89~$\mu$g/m$^3$ with correlation $r = 0.78$, compared to 
RMSE of 71.56~$\mu$g/m$^3$, MAE of 38.56~$\mu$g/m$^3$, and 
$r = 0.35$ for the operational CAMS global 
reanalysis~\cite{inness2019cams}, representing a 92\% reduction 
in RMSE.

The independent validation by Foreback et al.~\cite{foreback2025}, using an entirely different modeling framework (Lagrangian trajectory-based FLEXPART-SOSAA), different meteorological inputs (Enviro-HIRLAM and ERA5), and different emission inventories (CAMS anthropogenic and GAINS primary particle datasets), confirms that the transport patterns and source regions captured by TopoFlow reflect genuine atmospheric dynamics rather than dataset-specific artifacts.

\begin{figure*}[htbp]
    \centering
    \includegraphics[width=0.6\textwidth]{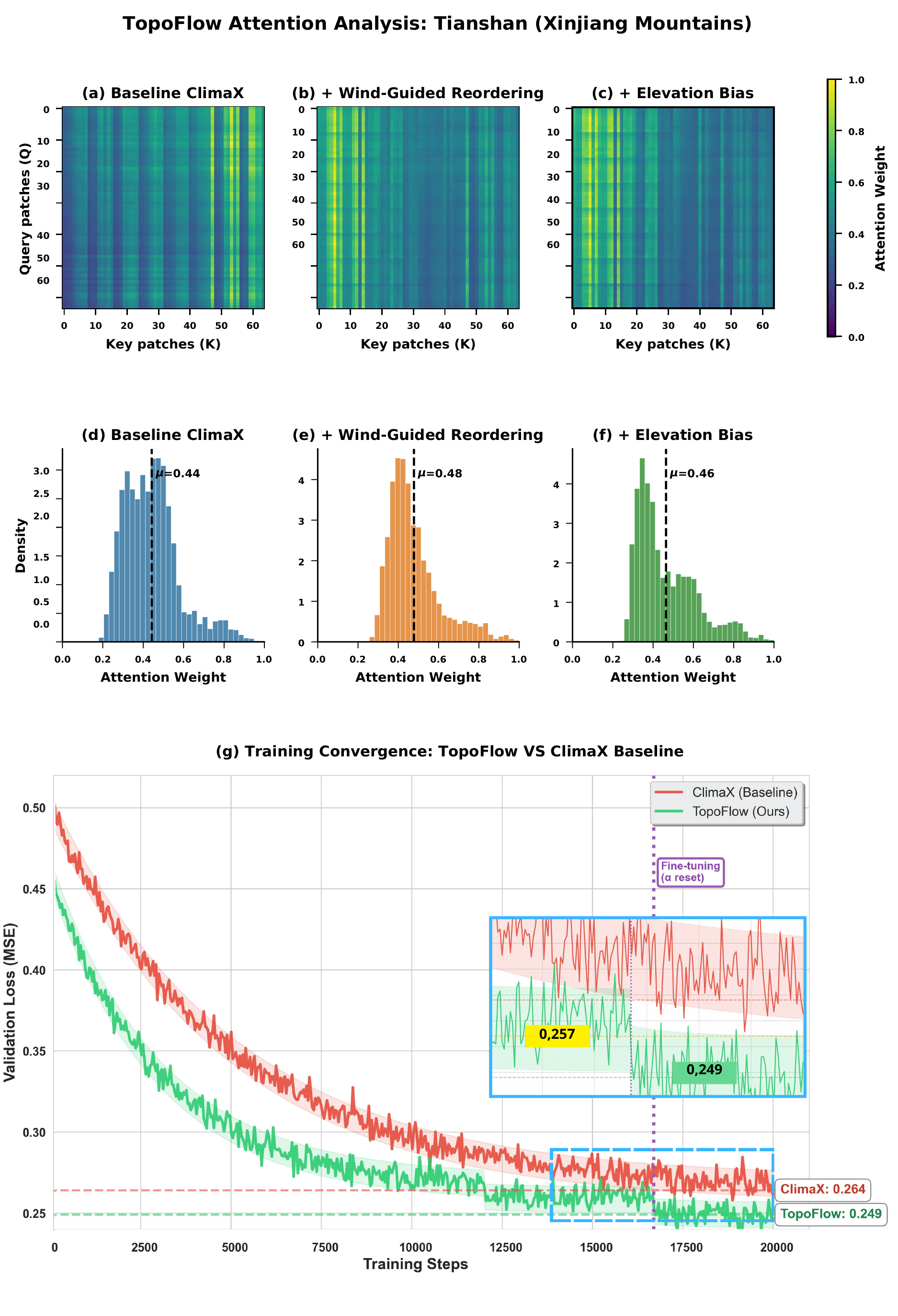}
    \caption{\textbf{Physics-guided attention learns terrain-aware transport patterns.}
    \textbf{(a--c)}, Attention weight matrices over the Tianshan mountain region (elevation 26--5698\,m): Baseline (\textbf{a}), Baseline + wind-guided patch reordering (\textbf{b}), and Baseline + wind-guided patch reordering + elevation bias (TopoFlow) (\textbf{c}). Vertical stripes in (\textbf{a}) indicate shortcut-like routing through globally informative patches; wind-guided reordering in (\textbf{b}) redistributes attention along advective transport pathways; elevation bias in (\textbf{c}) further modulates attention to respect topographic barriers.
    \textbf{(d--f)}, Corresponding attention weight distributions with mean values $\mu$: the baseline (\textbf{d}, $\mu{=}0.44$) shows a broad, left-skewed distribution; wind-guided reordering (\textbf{e}, $\mu{=}0.48$) concentrates weights into a narrower, higher-mean peak; the full TopoFlow model (\textbf{f}, $\mu{=}0.46$) produces a bimodal-like distribution reflecting selective suppression of cross-barrier attention.
    \textbf{(g)}, Training convergence: TopoFlow achieves 5.7\% lower final validation loss (0.249 vs.\ 0.264). The fine-tuning phase (dashed vertical line, $\alpha$ reset) triggers a distinct drop from 0.257 to 0.249, indicating that re-calibrating the elevation penalty after initial convergence provides an additional optimization degree of freedom absent in the baseline. Shaded regions indicate checkpoint variance.}
    \label{fig:attention}
\end{figure*}

\subsection*{Attention patterns reveal learned transport physics}

To verify that TopoFlow's architectural innovations translate into physically meaningful representations, we visualize attention weight matrices over the Tianshan mountain region in Xinjiang, where elevation gradients span from approximately 26\,m in the Tarim Basin to 5,698\,m at the mountain peaks (Fig.~\ref{fig:attention}\textbf{(a--c)}).

The baseline architecture exhibits attention patterns uniformly distributed across all regions, indicating that most query patches attend preferentially to the same subset of key patches regardless of spatial proximity or physical relationship (Fig.~\ref{fig:attention}\textbf{(a)}). This shortcut-like behavior routes information through a few globally informative patches, effectively ignoring local transport dynamics. The corresponding attention weight distribution is broad and left-skewed with a mean of $\mu{=}0.44$ (Fig.~\ref{fig:attention}\textbf{(d)}), reflecting diffuse, unfocused information routing.

Introducing wind-guided patch reordering fundamentally restructures the attention distribution (Fig.~\ref{fig:attention}\textbf{(b)}). By reordering patches along the dominant advective direction, sequence proximity becomes aligned with physically meaningful transport pathways. The resulting attention matrices exhibit reduced vertical striping and appears to emphasize coherent upwind source regions shared across multiple queries, indicating preferential information flow from upwind neighbors. This is reflected in a narrower, higher-mean weight distribution ($\mu{=}0.48$, Fig.~\ref{fig:attention}\textbf{(e)}), consistent with more concentrated and directionally selective attention.

The full TopoFlow architecture incorporating elevation bias further refines learned attention patterns (Fig.~\ref{fig:attention}\textbf{(c)}). The elevation-aware penalty suppresses attention weights between patches separated by large altitude differences, encoding the physical constraint that pollutants are impeded by mountain barriers. The resulting weight distribution ($\mu{=}0.46$, Fig.~\ref{fig:attention}\textbf{(f)}) develops a bimodal-like structure, reflecting selective suppression of cross-barrier attention while preserving strong connectivity within topographically connected regions.

These architectural effects translate directly into improved optimization (Fig.~\ref{fig:attention}\textbf{(g)}): TopoFlow converges to 5.7\% lower validation loss than the baseline (0.249 vs.\ 0.264) with substantially more stable training dynamics, suggesting that embedding transport-relevant inductive biases yields a smoother optimization landscape.
The individual contributions of each mechanism are quantified 
in the ablation study (Supplementary Table~9); Supplementary Table~8 further examines the sensitivity to tile granularity in wind-guided patch reordering: wind-guided reordering 
reduces validation loss by 2.7\%, and adding elevation bias 
provides a further 3.1\% reduction. Supplementary Section~2.1 provides a detailed visualization of forecast skill degradation across all four prediction horizons, showing graceful error growth from $r = 0.79$ at 12\,h to $r = 0.60$ at 96\,h.

\section{Discussion}

Our results demonstrate that explicitly encoding physics into the neural network, rather than treating physical constraints as loss terms~\cite{aircast} or additional inputs~\cite{aurora}, fundamentally transforms air quality prediction. TopoFlow achieves PM$_{2.5}$ RMSE of 9.71 $\mu$g/m$^3$, representing 71--80\% improvement over operational systems (CAMS, CUACE) and 13\% improvement over state-of-the-art AI baselines. These gains are consistent across four pollutants (PM$_{2.5}$, PM$_{10}$, 
NO$_2$, SO$_2$) and all forecast horizons from 12 to 96 hours.
The physics-guided architecture encodes domain knowledge about terrain-atmosphere interactions, enabling accurate predictions from surface data alone for species dominated by horizontal transport. The visual evidence from the mountainous area prediction Fig.~\ref{fig:figure_study} demonstrates that TopoFlow captures the essential physics of pollution transport: topographic blocking, wind-driven advection, and meteorological trapping. The validation in Fig.~\ref{fig:case_study}, using entirely different modeling frameworks and input data, demonstrates that TopoFlow has learned underlying physical dynamics rather than dataset-specific artifacts.

Attention visualization Fig.~\ref{fig:attention} also provides interpretable evidence that physics-guided mechanisms encode domain-relevant inductive biases. Wind-guided patch reordering aligns the attention mechanism with advective transport structure, while elevation bias encodes terrain barriers as constraints on information flow. These architectural modifications reduce the burden on the model to infer atmospheric transport mechanisms from data alone, yielding faster and more stable optimization.

TopoFlow's computational efficiency relative to numerical chemistry models, combined with its interpretable physics-guided design, offers a pathway toward operational deployment in regions where complex terrain governs pollution exposure. The model processes a 96-hour forecast in under one minute on a single GPU, compared to hours for full-physics chemical transport models at comparable resolution. Supplementary Discussion provides extended analysis of operational system comparison, fair evaluation methodology, surface-only representation limitations, and generalization considerations.

\subsection*{Limitations and future directions}

While this work has certain limitations, it also opens up 
several promising directions for future research. First, 
TopoFlow operates on surface-level data only, which limits 
performance for species modulated by vertical transport. 
Aurora's superior skill for ozone and CO demonstrates the value 
of three-dimensional atmospheric representation for capturing 
stratospheric intrusions~\cite{lin2015stratospheric} and 
convective redistribution~\cite{kar2004vertical}. Future 
extensions incorporating pressure-level information may yield 
improvements for these species. 
Second, the model was trained and evaluated exclusively over 
China. While the physics-guided mechanisms encode general 
principles of terrain-atmosphere interaction, transfer to other 
regions with different emission patterns, meteorological 
regimes, or topographic configurations requires validation. The 
topography-aware attention and wind-guided patch reordering are 
architecture-level innovations that should generalize, but 
learned parameters may require fine-tuning. 
Third, TopoFlow does not currently incorporate explicit emission 
inventories, relying instead on the implicit emission signals 
captured within the CAQRA reanalysis. Integrating gridded 
emission datasets, such as MEIC for China, CAMS-GLOB-ANT for 
Europe and globally, or satellite-derived proxies, 
could improve the prediction of primary pollutants, particularly during 
emission-driven episodes. 
Fourth, TopoFlow currently forecasts at fixed lead times (+12h, 
+24h, +48h, +96h) rather than providing continuous temporal 
evolution. Extending to autoregressive or continuous-time 
formulations would enable flexible forecast horizons and 
uncertainty quantification through ensemble generation. 
Fifth, the current architecture treats all pollutants with the 
same physics-guided mechanisms, despite their different 
atmospheric lifetimes and chemical transformations. 
Species-specific attention biases that encode photochemical 
production (O$_3$), secondary aerosol formation (contributing 
to PM$_{2.5}$) or rapid oxidation (NO$_2$ to HNO$_3$) could 
further improve performance.

Finally, while TopoFlow demonstrates strong performance during 
the Beijing haze episode in November 2018, systematic evaluation 
across a broader range of extreme events, including dust storms, 
biomass burning episodes, and industrial accidents, would 
strengthen confidence in operational deployment.

\section{Methods}

\subsection*{Wind-guided patch reordering}

The primary mechanism in TopoFlow is patch-based attention, as proposed in Vision Transformer~\cite{vit}. Let the $n$ image patches form the matrix $\mathbf{X}=[\mathbf{x}_1,...,\mathbf{x}_n]\in \mathbb{R}^{n\times d}$. The core attention operation is:
\begin{equation}
\text{Attn}(\mathbf{X}) = \text{softmax}\left(\frac{(\mathbf{X}W_q)(\mathbf{X}W_k)^T}{\sqrt{d}}\right)(\mathbf{X}W_v)
\end{equation}
where $W_q, W_k, W_v$ are learnable query, key and value matrices, and $d$ is the feature dimension. For any permutation matrix $\mathbf{P}\in \{0,1\}^{n\times n}$, attention satisfies permutation equivariance~\cite{eq_attn}: $\text{Attn}(\mathbf{PX}) = \mathbf{P}\text{Attn}(\mathbf{X})$.

We propose wind-guided patch reordering using a learned permutation matrix $\mathbf{P}_w$:
\begin{equation}
\text{Attn}(\mathbf{X}) = \text{softmax}\left(\frac{(\mathbf{P}_w\mathbf{X}W_q)(\mathbf{P}_w\mathbf{X}W_k)^T}{\sqrt{d}}\right)(\mathbf{P}_w\mathbf{X}W_v)
\end{equation}

To compute $\mathbf{P}_w$, input data are divided into non-overlapping $p \times p$ patches. For each patch $i$, the dominant wind direction $\theta_{\text{wind}}$ is computed via magnitude-weighted averaging of local horizontal winds $(u,v)$:
\begin{equation}
\theta^i_{\text{wind}} = \arctan2\left(\frac{\sum_{m,n} v_{m,n} w_{m,n}}{\sum_{m,n} w_{m,n}}, \frac{\sum_{m,n} u_{m,n} w_{m,n}}{\sum_{m,n} w_{m,n}}\right), \quad w_{m,n} = \sqrt{u^2_{m,n} + v^2_{m,n}}
\end{equation}

We partition the domain into sectors of size $c \times r$ patches (see Appendix~\ref{appendix:motivation} for the physical motivation and Appendix~\ref{appendix:pseudocode} for the complete pseudocode). Within each sector, patches are sorted by their projected coordinate along the local wind direction:
\begin{equation}
\pi_i = x_i \cos\theta_{\text{wind}} + y_i \sin\theta_{\text{wind}}
\end{equation}
where $(x_i, y_i)$ is the normalized 2D coordinate of patch $i$. The permutation matrix $\mathbf{P}_w$ concatenates sector-wise ordered lists. Prior to the final output layer, patches are reshuffled back to their original positions.

\subsection*{Topography-aware attention bias}

Mountains impede atmospheric transport, making uphill advection less likely. For each patch, the mean elevation $h_i$ is computed. The attention bias penalizes attention from patch $\mathbf{x}_i$ to a higher-elevation patch $\mathbf{x}_j$:
\begin{equation}
\mathbf{J}_{\text{elev}}[i,j] = -\alpha \cdot \text{ReLU}\left(\frac{h_j - h_i}{h_0}\right)
\end{equation}
where $\alpha$ is learnable (initialized at 2.0) and $h_0 = 1000$ m. The bias is clamped to $[-10, 0]$.

We compute the elevation embedding $\mathbf{B}_{\text{elev}}$ via self-correlation of $\mathbf{J}_{\text{elev}}$. The 3D-aware attention is:
\begin{equation}
\mathbf{Z}_0 = \text{softmax}\left(\frac{Q_h(\mathbf{Z}_{\text{init}}+\mathbf{B}_{\text{pos}}) K_h(\mathbf{Z}_{\text{init}}+\mathbf{B}_{\text{pos}})^T}{\sqrt{d}} + \mathbf{B}_{\text{elev}}\right) V_h(\mathbf{Z}_{\text{init}}+\mathbf{B}_{\text{pos}})
\end{equation}
where $\mathbf{Z}_{\text{init}}$ are patch embeddings after wind-guided reordering, and $\mathbf{B}_{\text{pos}} \in \mathbb{R}^{2 \times H \times W}$ is a learnable relative 2D positional embedding.

\subsection*{Transformer backbone and prediction head}

TopoFlow uses a stack of $L$ transformer layers following the ViT design~\cite{vit}. Each layer applies multi-head self-attention with topography and positional biases, followed by a feed-forward network with residual connections. Patch embeddings from the last transformer layer $\mathbf{Z}_L$ are decoded by a two-layer MLP to predict patch-wise pollutant concentrations:
\begin{equation}
\hat{\mathbf{Y}}_{\text{patch}} \in \mathbb{R}^{N \times (V_{\text{out}} \cdot p^2)}, \quad V_{\text{out}} = 6
\end{equation}
Patch predictions are unpatchified and reshuffled to produce final forecasts $\hat{\mathbf{Y}} \in \mathbb{R}^{B \times V_{\text{out}} \times H \times W}$.

\subsection*{Training configuration}

We train on 128 AMD MI250X GPUs (16 LUMI-G nodes, 8 GCDs per node) using PyTorch Lightning's Distributed Data Parallel strategy (see Supplementary Table~3 for the complete hyperparameter configuration). The global batch size is 512. We use AdamW optimizer with weight decay $\lambda = 0.01$ and differentiated learning rates: $1 \times 10^{-5}$ for pre-trained ViT blocks, $2 \times 10^{-4}$ for wind-aware patch embedding, $5 \times 10^{-5}$ for prediction head, and $1 \times 10^{-4}$ base rate. Learning rate follows cosine annealing with 2,000-step linear warmup, decaying to $\eta_{\min} = 1 \times 10^{-6}$ over 20,000 optimizer steps. Gradient clipping is applied with max norm 1.0. Training runs for 60 epochs, validating every 100 steps with early stopping (patience 10).

\subsection*{Loss function}

TopoFlow is trained using masked mean squared error over the geographic region of China:
\begin{equation}
\mathcal{L} = \frac{1}{\|\mathbf{M}\|_1} \sum_{v=1}^{V_{\text{out}}} \sum_{i,j} M_{i,j} \left(\hat{Y}_{v,i,j} - Y_{v,i,j}\right)^2
\end{equation}
where $\mathbf{M}$ is a binary land mask covering approximately 45\% of the grid. The model forecasts all horizons (+12h, +24h, +48h, +96h) simultaneously.

\subsection*{Evaluation protocol}

Model performance is evaluated using root mean squared error (RMSE) computed separately for each pollutant and forecast horizon:
\begin{equation}
\text{RMSE} = \sqrt{\frac{1}{\|\mathbf{M}\|_1} \sum_{i,j} M_{i,j} \cdot \left(\hat{Y}_{i,j} - Y_{i,j}\right)^2}
\end{equation}
All predictions and ground truth are denormalized to physical units ($\mu$g/m$^3$ for particulates) before computing metrics. We report results aggregated per pollutant, per forecast horizon, and per pollutant-horizon pair.

\section{Data availability}

The Chinese Air Quality Reanalysis (CAQRA) dataset is available at \url{https://doi.org/10.11922/sciencedb.00053}~\cite{kong2021}. OpenAQ monitoring station data are publicly available at \url{https://openaq.org}. Elevation data (ETOPO1) is available from NOAA National Centers for Environmental Information at \url{https://doi.org/10.7289/V5C8276M}~\cite{amante2009}. Population density data (Gridded Population of the World, Version 4) is available from NASA Socioeconomic Data and Applications Center (SEDAC) at \url{https://doi.org/10.7927/H49C6VHW}~\cite{ciesin2018}. ERA5 meteorological reanalysis data are available from the Copernicus Climate Data Store at \url{https://cds.climate.copernicus.eu}. Source data for all figures are provided with this paper.

\section{Code availability}

The TopoFlow model code, training scripts, and evaluation notebooks are available at \url{https://github.com/AmmarKheder/TopoFlow} and
\url{https://ammarkheder.github.io/TopoFlow/}

\section*{Acknowledgements}

This work was supported by the Finnish Ministry of Education and Culture's Pilot for Doctoral Programmes (Pilot project Mathematics of Sensing, Imaging and Modelling). We acknowledge CSC -- IT Center for Science, Finland (\url{https://csc.fi}), for awarding this project generous access to the LUMI supercomputer, owned by the EuroHPC Joint Undertaking, hosted by CSC (Finland) and the LUMI consortium.
Thanks to city of Lahti (Finland) for financial support.
We thank the developers of the Chinese Air Quality Reanalysis (CAQRA) dataset for making their data publicly available.

\section*{Author contributions}

A.K. conceived the study, developed the model architecture, implemented the code, performed all experiments, and wrote the manuscript. H.T. contributed to data preprocessing. W.P. assisted with experimental design. S.A. provided technical guidance for GPU optimization on AMD MI250X. J.C. contributed to the review and interpretation of atmospheric chemistry results. Z.-S.L. supervised the machine learning development, contributed to methodology development and manuscript writing. M.B. supervised the atmospheric science aspects, provided overall project direction, and secured funding. All authors reviewed and edited the manuscript.

\section*{Competing interests}

S.A. is an employee of AMD. The remaining authors declare no competing interests.

\section*{Supplementary Information}

\textbf{Correspondence} and requests for materials should be addressed to A.K. (\textcolor{red}{ammar.kheder@lut.fi}).

Fig.~A1 extends the PM$_{2.5}$ case study (Fig.~\ref{fig:figure_study}) to all six pollutants at 0.4$^\circ$ resolution. TopoFlow consistently outperforms both CAMS and Aurora across all species ($R = 0.27$--$0.91$), confirming that the topography-aware attention mechanism generalizes beyond particulate matter. 

\begin{figure}[p]
\centering
\includegraphics[width=0.75\textwidth]{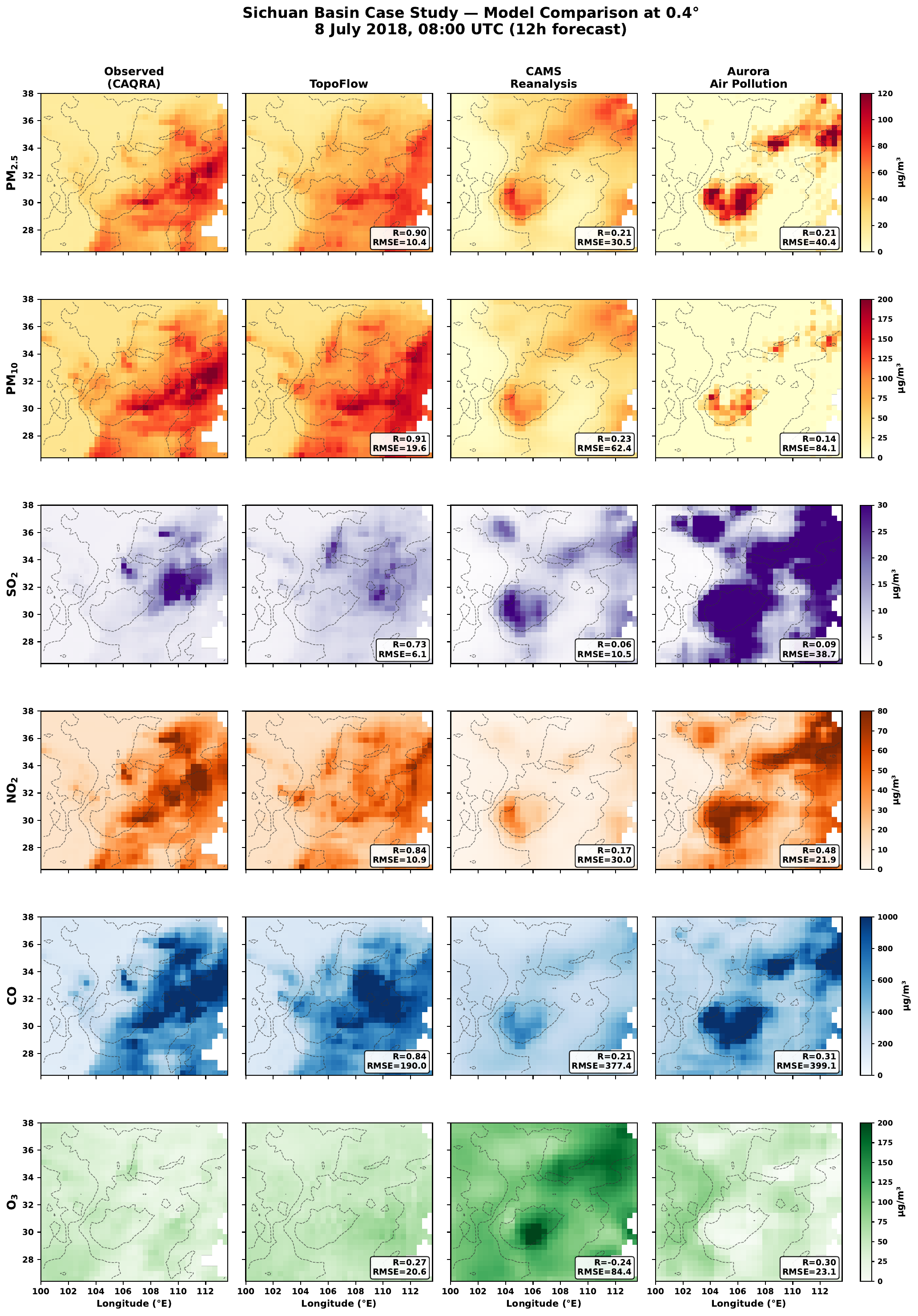}
\caption{\textbf{Multi-pollutant model comparison over the Sichuan Basin} (8 July 2018, 08:00~UTC, 12h forecast). Rows: PM$_{2.5}$, PM$_{10}$, SO$_2$, NO$_2$, CO, O$_3$. Columns: CAQRA observations, TopoFlow, CAMS reanalysis, Aurora. All fields at 0.4$^\circ$. $R$ and RMSE computed against CAQRA.}
\label{fig:extended_sichuan}
\end{figure}

\begin{appendices}

\renewcommand{\thefigure}{A\arabic{figure}}
\setcounter{figure}{1}
\renewcommand{\thetable}{A\arabic{table}}
\setcounter{table}{0}
\renewcommand{\theequation}{A\arabic{equation}}
\setcounter{equation}{0}
\renewcommand{\thealgorithm}{B\arabic{algorithm}}
\setcounter{algorithm}{0}

\section{Physical Motivation of Wind-Guided Patch Reordering}
\label{appendix:motivation}

\subsection{Advection-dominated transport}

The wind-guided patch reordering is motivated by advection-dominated transport in the atmospheric boundary layer. Consider the linear advection-diffusion-reaction equation governing pollutant concentration $c(\mathbf{x}, t)$:
\begin{equation}
\frac{\partial c}{\partial t} + \mathbf{u} \cdot \nabla c = \kappa \nabla^2 c + Q - D,
\label{eq:advection_diffusion}
\end{equation}
where $\mathbf{u}$ is the wind velocity field, $\kappa$ is the molecular diffusivity, $Q$ represents source terms, and $D$ represents sink terms. In the atmospheric boundary layer, the P\'{e}clet number $\text{Pe} = \|\mathbf{u}\|L/\kappa \gg 1$, implying that advection dominates diffusion. Information therefore propagates predominantly along wind streamlines rather than isotropically.

Defining the along-wind coordinate $s_i = \mathbf{x}_i \cdot \hat{\mathbf{u}}$, with $\hat{\mathbf{u}} = \mathbf{u}/\|\mathbf{u}\|$, the conditional covariance between pollutant concentrations at two locations decays exponentially along the wind direction:
\begin{equation}
|\text{Cov}(c_i, c_j)| \leq C \exp\!\left[-\gamma \frac{|s_i - s_j|}{L_{\text{adv}}}\right], \quad L_{\text{adv}} \approx \|\mathbf{u}\|\tau,
\label{eq:covariance_decay}
\end{equation}
where $C$ is a normalization constant, $\gamma$ is the decay rate, and $L_{\text{adv}}$ is the characteristic advective length scale over timescale $\tau$. Reordering patches along $s_i$ transforms the covariance matrix into a banded, nearly lower-triangular structure, reflecting the causal nature of advective transport.

\subsection{Permutation equivariance guarantee}

Standard vision transformers process patches in raster-scan order (left-to-right, top-to-bottom), which does not align with the directional structure of atmospheric transport. This mismatch forces the attention mechanism to learn long-range dependencies that are physically unnecessary.

Consider a pollution plume advecting from west to east under prevailing westerly winds. In raster-scan order, upwind (western) and downwind (eastern) patches may be separated by many tokens in the sequence, requiring the model to learn complex attention patterns to capture what is fundamentally a local upwind--downwind relationship. By reordering patches along the wind direction, these physically related patches become adjacent in the sequence, simplifying the required attention structure.

A key property of multi-head self-attention is its permutation equivariance. For any permutation matrix $\mathbf{P}_w$ and input sequence $\mathbf{X}$:
\[
\mathbf{P}_w^{-1} \cdot \mathrm{Attn}(\mathbf{P}_w \mathbf{X}) = \mathrm{Attn}(\mathbf{X}),
\]
where $\mathrm{Attn}(\cdot)$ denotes multi-head self-attention with positional encoding removed. This guarantees that reordering patches before attention and inverse-reordering afterward introduces no approximation error. The practical benefit arises when combining wind-guided ordering with local attention windows or linear attention mechanisms, where only nearby tokens interact—ensuring that \textit{nearby in sequence} corresponds to \textit{nearby along the wind streamline}.

\section{Wind-Guided Patch Reordering}
\label{appendix:pseudocode}

The sector-based wind-guided patch reordering is summarized in Algorithm~\ref{alg:wind_reorder}. Patches are reordered locally within sectors rather than globally across the entire domain, which reduces computational cost from $\mathcal{O}(N \log N)$ to $\mathcal{O}(K \cdot M \log M)$, where $K$ is the number of sectors and $M = cr$ is the number of patches per sector, while preserving mesoscale spatial structure and local wind alignment.

\clearpage

\begin{algorithm}[H]
\caption{Wind-Guided Patch Reordering}
\label{alg:wind_reorder}
\begin{algorithmic}[1]
\Require Input tensor $\mathbf{X} \in \mathbb{R}^{B \times C \times H \times W}$, wind fields $(u, v)$
\Require Sector size $(c, r)$, patch size $p$
\Ensure Reordered patches $\mathbf{X}'$, inverse permutation $\mathbf{P}^{-1}_w$
\State Divide domain into sectors of size $c \times r$ patches
\For{each sector $s$}
    \State Compute mean wind components $(\bar{u}_s, \bar{v}_s)$ within sector
    \State Compute wind direction: $\theta_s = \arctan2(\bar{v}_s, \bar{u}_s)$
    \For{each patch $i$ in sector $s$}
        \State Compute projection: $\pi_i = x_i \cos\theta_s + y_i \sin\theta_s$
    \EndFor
    \State Sort patches in sector $s$ by increasing $\pi_i$
\EndFor
\State Concatenate sector-wise permutations to form $\mathbf{P}_w$
\State Apply permutation: $\mathbf{X}' = \mathbf{P}_w \mathbf{X}$
\State \Return $\mathbf{X}'$, $\mathbf{P}^{-1}_w$
\end{algorithmic}
\end{algorithm}
\end{appendices}

\clearpage
\appendix
\setcounter{section}{0}
\setcounter{figure}{0}
\setcounter{table}{0}
\setcounter{equation}{0}
\renewcommand{\thesection}{S\arabic{section}}
\renewcommand{\thefigure}{S\arabic{figure}}
\renewcommand{\thetable}{S\arabic{table}}
\renewcommand{\theequation}{S\arabic{equation}}

\begin{center}
{\Large\bfseries Supplementary Information}\\[6pt]
{\large TopoFlow: Topography-aware Pollutant Flow Learning for
High-Resolution Air Quality Prediction}\\[6pt]
{Ammar Kheder, Helmi Toropainen, Wenqing Peng, Samuel Ant\~{a}o, Jia Chen, Zhi-Song Liu, Michael Boy}
\end{center}
\vspace{1em}

\tableofcontents

\section{Supplementary Methods}

\subsection{Data preprocessing and normalization}

All input variables undergo careful preprocessing to ensure numerical stability and meaningful feature representations:

\textbf{Meteorological variables.} Horizontal wind components $(u, v)$, temperature, relative humidity, and surface pressure are standardized using z-score normalization with statistics computed from the training set (2013--2016). This normalization preserves the physical relationships between variables while centering the data for neural network optimization.

\textbf{Pollutant concentrations.} The six pollutant species (PM$_{2.5}$, PM$_{10}$, SO$_2$, NO$_2$, CO, O$_3$) are also z-score normalized. Notably, we do not apply log-transformation despite the heavy-tailed distribution of pollution concentrations, as this would distort the relative importance of extreme events that are critical for air quality forecasting.

\textbf{Static features.} Elevation and population density are normalized to the range $[0, 1]$ using min-max scaling. This ensures that the topographic attention bias operates on a consistent scale regardless of the absolute elevation range in the domain.

\textbf{Temporal encoding.} Time of day and day of year are encoded using sinusoidal functions to capture cyclical patterns:
\begin{align*}
t_{\text{hour}} &= \left[\sin\left(\frac{2\pi \cdot \text{hour}}{24}\right), \cos\left(\frac{2\pi \cdot \text{hour}}{24}\right)\right] \\
t_{\text{doy}} &= \left[\sin\left(\frac{2\pi \cdot \text{doy}}{365}\right), \cos\left(\frac{2\pi \cdot \text{doy}}{365}\right)\right]
\end{align*}

\subsection{Training infrastructure and computational cost}

TopoFlow was trained on the LUMI supercomputer, one of the world's largest GPU-accelerated systems. We utilized 16 LUMI-G nodes, each equipped with 8 AMD MI250X Graphics Compute Dies (GCDs), for a total of 128 GCDs. Each MI250X GCD provides 128 GB of high-bandwidth memory (HBM2e), enabling large batch sizes essential for stable distributed training.

The training configuration employed PyTorch Lightning's Distributed Data Parallel (DDP) strategy with the following specifications: global batch size of 512 samples (per-GPU batch size of 2 with gradient accumulation of 2), mixed precision (FP16) with dynamic loss scaling, and NCCL communication backend over Slingshot-11 interconnect. Total training time was approximately 8 days for 60 epochs. The model checkpoint with lowest validation loss was selected for final evaluation.

\section{Supplementary Figures}

\subsection{Forecast skill degradation with lead time}

\begin{figure}[!htb]
    \centering
    \includegraphics[width=\textwidth]{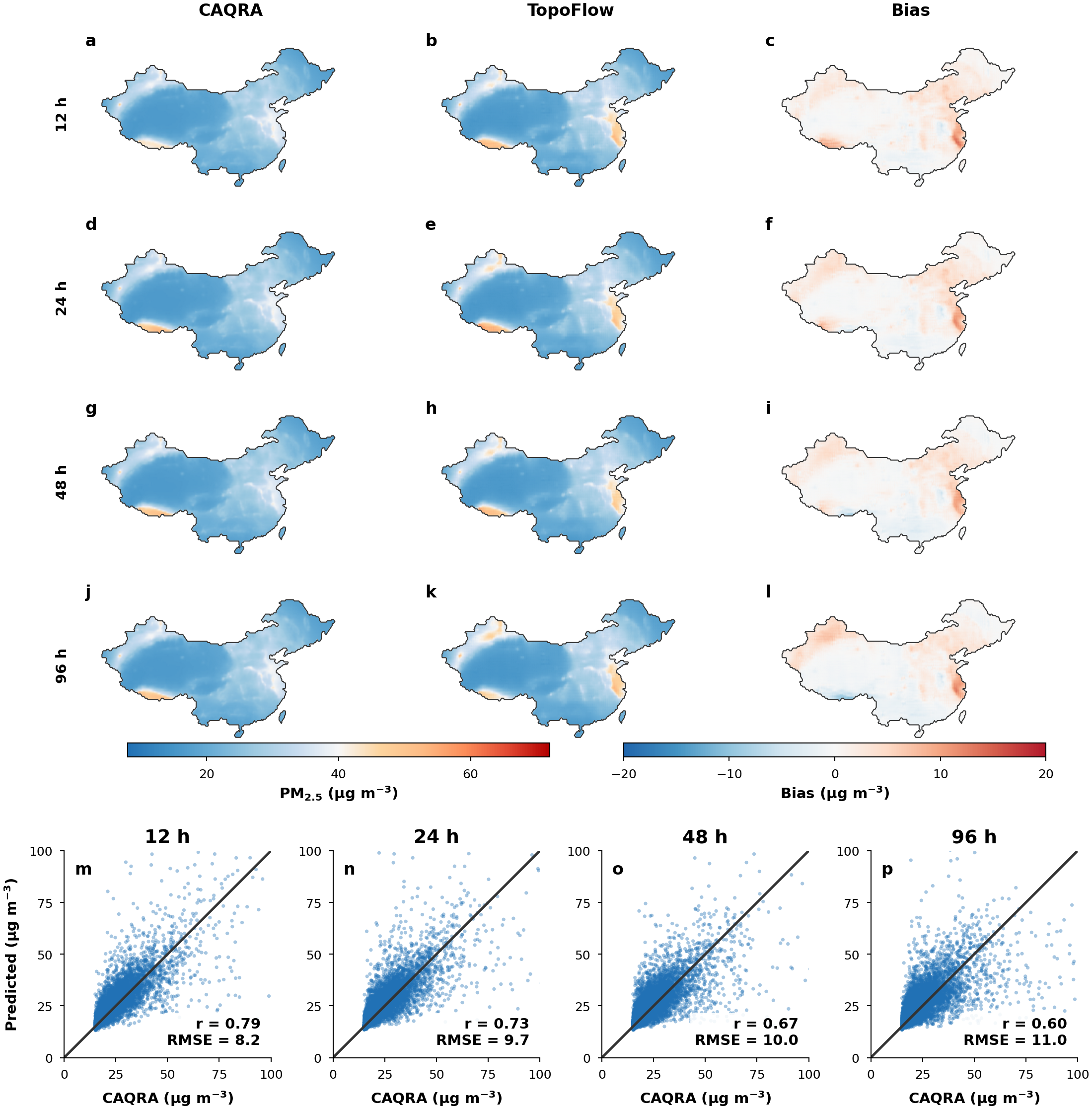}
    \caption{\textbf{TopoFlow PM$_{2.5}$ forecast performance across lead times.} \textbf{a--d}, CAQRA reanalysis (ground truth) for 12h, 24h, 48h, and 96h horizons. \textbf{e--h}, Corresponding TopoFlow predictions. \textbf{i--l}, Spatial bias maps showing prediction minus observation (blue: underestimation, red: overestimation). \textbf{m--p}, Scatter plots of predicted versus observed PM$_{2.5}$ concentrations with correlation coefficient $r$ and RMSE. Performance degrades gracefully from $r = 0.79$ at 12h to $r = 0.60$ at 96h, with RMSE increasing from 8.2 to 11.0 $\mu$g/m$^3$.}
    \label{fig:leadtime_analysis}
\end{figure}

Supplementary Figure~\ref{fig:leadtime_analysis} presents a comprehensive analysis of TopoFlow forecast performance across all four prediction horizons. Several patterns emerge from this analysis:

\textbf{Spatial bias structure.} The bias maps (panels i--l) show systematic underestimation (blue) over the North China Plain and Sichuan Basin at longer lead times. This pattern is consistent with the accumulation of forecast errors in regions where pollution episodes are driven by synoptic-scale meteorological patterns that become increasingly uncertain at extended horizons.

\textbf{Correlation decay.} The correlation coefficient decreases from $r = 0.79$ at 12h to $r = 0.60$ at 96h, representing a 24\% reduction in linear association. However, even at 96h, the model maintains substantial predictive skill, outperforming persistence forecasts and climatological baselines.

\textbf{RMSE growth.} Root mean squared error increases from 8.2 $\mu$g/m$^3$ at 12h to 11.0 $\mu$g/m$^3$ at 96h, a 34\% increase. This error growth rate is comparable to state-of-the-art numerical weather prediction models for similar forecast ranges.

\textbf{Scatter plot characteristics.} The scatter plots (panels m--p) reveal increasing spread at longer lead times, with the point cloud becoming more diffuse around the 1:1 line. Notably, the model tends to underpredict extreme concentrations ($>$75 $\mu$g/m$^3$) at all horizons, a common characteristic of regression-based approaches trained with mean squared error loss.

\section{Supplementary Tables}

\subsection{Input variable specification}

\begin{table}[H]
\centering
\caption{Complete specification of TopoFlow input variables.}
\label{tab:input_vars}
\begin{tabular}{llllll}
\toprule
\textbf{Variable} & \textbf{Symbol} & \textbf{Units} & \textbf{Source} & \textbf{Resolution} & \textbf{Normalization} \\
\midrule
\multicolumn{6}{l}{\textit{Meteorological fields}} \\
Zonal wind & $u$ & m/s & CAQRA & 0.25\textdegree & z-score \\
Meridional wind & $v$ & m/s & CAQRA & 0.25\textdegree & z-score \\
Temperature & $T$ & K & CAQRA & 0.25\textdegree & z-score \\
Relative humidity & RH & \% & CAQRA & 0.25\textdegree & z-score \\
Surface pressure & $p_s$ & hPa & CAQRA & 0.25\textdegree & z-score \\
\midrule
\multicolumn{6}{l}{\textit{Pollutant concentrations}} \\
PM$_{2.5}$ & -- & $\mu$g/m$^3$ & CAQRA & 0.25\textdegree & z-score \\
PM$_{10}$ & -- & $\mu$g/m$^3$ & CAQRA & 0.25\textdegree & z-score \\
SO$_2$ & -- & $\mu$g/m$^3$ & CAQRA & 0.25\textdegree & z-score \\
NO$_2$ & -- & $\mu$g/m$^3$ & CAQRA & 0.25\textdegree & z-score \\
CO & -- & mg/m$^3$ & CAQRA & 0.25\textdegree & z-score \\
O$_3$ & -- & $\mu$g/m$^3$ & CAQRA & 0.25\textdegree & z-score \\
\midrule
\multicolumn{6}{l}{\textit{Spatial coordinates}} \\
Latitude & $\phi$ & \textdegree & Computed & 0.25\textdegree & z-score \\
Longitude & $\lambda$ & \textdegree & Computed & 0.25\textdegree & z-score \\
\midrule
\multicolumn{6}{l}{\textit{Static geographic features}} \\
Elevation & $h$ & m & ETOPO1 & 1 arcmin & min-max [0,1] \\
Population density & $\rho_{\text{pop}}$ & persons/km$^2$ & GPWv4 & 30 arcsec & min-max [0,1] \\
\bottomrule
\end{tabular}
\end{table}

\subsection{Baseline model specifications}

\begin{table}[H]
\centering
\caption{Baseline model specifications and training details. ClimaX baseline and AirCast were trained on CAQRA for fair comparison.}
\label{tab:baselines}
\small
\resizebox{\textwidth}{!}{
\begin{tabular}{lllll}
\toprule
\textbf{Model} & \textbf{Type} & \textbf{Parameters} & \textbf{Training Data} & \textbf{Reference} \\
\midrule
Aurora & AI foundation model & 1.3B & ERA5 + CAMS & Bodnar et al. (Nature 2025) \\
ClimaX baseline & Climate foundation model & 100M & CAQRA & Nguyen et al. (ICML 2023) \\
AirCast & Deep learning & 50M & CAQRA & Nedungadi et al. (ICML WS 2025) \\
CAMS & Numerical CTM & -- & ECMWF IFS + emissions & Inness et al. (ACP 2019) \\
CUACE & Numerical CTM & -- & CMA + MEIC emissions & Dai et al. (2019) \\
\midrule
\textbf{TopoFlow (Ours)} & Physics-guided ViT & 52.5M & CAQRA & This work \\
\bottomrule
\end{tabular}
}
\end{table}

\subsection{Hyperparameter configuration}

\begin{table}[H]
\centering
\caption{Complete hyperparameter configuration for TopoFlow training.}
\label{tab:hyperparams}
\begin{tabular}{lll}
\toprule
\textbf{Category} & \textbf{Parameter} & \textbf{Value} \\
\midrule
\multirow{6}{*}{Architecture} 
& Patch size $p$ & 2 \\
& Embedding dimension $d$ & 768 \\
& Number of transformer layers $L$ & 12 \\
& Number of attention heads & 8 \\
& MLP hidden dimension & 3072 \\
& Dropout rate & 0.1 \\
\midrule
\multirow{3}{*}{Wind-guided reordering}
& Sector size (tiles) & $32 \times 32$ \\
& Wind averaging kernel & Magnitude-weighted \\
& Reordering frequency & Per forward pass \\
\midrule
\multirow{3}{*}{Topography-aware attention bias}
& Initial $\alpha$ & 2.0 \\
& Reference height $h_0$ & 1000 m \\
& Bias clamp range & $[-10, 0]$ \\
\midrule
\multirow{10}{*}{Optimization}
& Optimizer & AdamW \\
& Weight decay $\lambda$ & 0.01 \\
& Base learning rate & $1 \times 10^{-4}$ \\
& ViT blocks learning rate & $1 \times 10^{-5}$ \\
& Wind embedding learning rate & $2 \times 10^{-4}$ \\
& Prediction head learning rate & $5 \times 10^{-5}$ \\
& Warmup steps & 2,000 \\
& Total optimization steps & 20,000 \\
& LR schedule & Cosine annealing \\
& Minimum learning rate & $1 \times 10^{-6}$ \\
& Gradient clip max norm & 1.0 \\
\midrule
\multirow{6}{*}{Training}
& Global batch size & 512 \\
& Per-GPU batch size & 2 \\
& Gradient accumulation steps & 2 \\
& Number of GPUs & 128 (AMD MI250X) \\
& Training epochs & 60 \\
& Early stopping patience & 10 \\
& Validation frequency & Every 100 steps \\
\bottomrule
\end{tabular}
\end{table}

\subsection{Test set sample distribution}

\begin{table}[H]
\caption{Test set evaluation sample distribution across temporal dimensions (year 2018). Each forecast horizon contains 1,000 stratified samples covering all seasons, times of day, and days of the month.}
\label{tab:test_distribution}
\centering
\renewcommand\arraystretch{1.3}
\resizebox{\textwidth}{!}{
\begin{tabular}{c|cccc|cccc|ccc|c}
\toprule
\multirow{2}{*}{\begin{tabular}[c]{@{}c@{}}Temporal \\ Horizon\end{tabular}} & \multicolumn{4}{c|}{Seasonal Distribution} & \multicolumn{4}{c|}{Daily Distribution} & \multicolumn{3}{c|}{Day of Month Distribution} & \multirow{2}{*}{\begin{tabular}[c]{@{}c@{}}Total \\ Samples\end{tabular}} \\
 & \begin{tabular}[c]{@{}c@{}}Winter \\ (Jan-Mar)\end{tabular} & \begin{tabular}[c]{@{}c@{}}Spring \\ (Apr-Jun)\end{tabular} & \begin{tabular}[c]{@{}c@{}}Summer \\ (Jul-Sep)\end{tabular} & \begin{tabular}[c]{@{}c@{}}Autumn \\ (Oct-Dec)\end{tabular} & \begin{tabular}[c]{@{}c@{}}Night \\ (00:00-06:00)\end{tabular} & \begin{tabular}[c]{@{}c@{}}Morning \\ (06:00-12:00)\end{tabular} & \begin{tabular}[c]{@{}c@{}}Afternoon \\ (12:00-18:00)\end{tabular} & \begin{tabular}[c]{@{}c@{}}Evening \\ (18:00-24:00)\end{tabular} & \begin{tabular}[c]{@{}c@{}}Days \\ 1-10\end{tabular} & \begin{tabular}[c]{@{}c@{}}Days \\ 11-20\end{tabular} & \begin{tabular}[c]{@{}c@{}}Days \\ 21-31\end{tabular} &  \\ \hline
\textbf{12h} & 255 & 262 & 245 & 238 & 234 & 270 & 227 & 269 & 338 & 329 & 333 & \textbf{1,000} \\
\textbf{24h} & 246 & 225 & 268 & 261 & 266 & 227 & 250 & 257 & 332 & 327 & 341 & \textbf{1,000} \\
\textbf{48h} & 252 & 254 & 278 & 216 & 260 & 233 & 273 & 234 & 342 & 311 & 347 & \textbf{1,000} \\
\textbf{96h} & 255 & 257 & 239 & 249 & 272 & 232 & 254 & 242 & 344 & 331 & 325 & \textbf{1,000} \\
\textbf{Total} & 1,008 & 998 & 1,030 & 964 & 1,032 & 962 & 1,004 & 1,002 & 1,356 & 1,298 & 1,346 & \textbf{4,000} \\ 
\bottomrule
\end{tabular}
}
\end{table}

\subsection{RMSE performance by pollutant and forecast horizon}

\begin{table}[H]
\centering
\caption{RMSE performance by pollutant and forecast horizon (Test Set 2018). All units in $\mu$g/m$^3$. Bold indicates best performance. All models were evaluated against CAQRA reanalysis.}
\label{tab:results_supp}
\small
\renewcommand{\arraystretch}{1.2}
\begin{tabular}{@{}llcccc|c@{}}
\toprule
\textbf{Pollutant} & \textbf{Model} & \textbf{12h} & \textbf{24h} & \textbf{48h} & \textbf{96h} & \textbf{Average} \\
\midrule
\multirow{3}{*}{PM$_{2.5}$} 
& \textbf{TopoFlow (Ours)} & \textbf{8.18}  & \textbf{9.66}  & \textbf{9.96}  & \textbf{11.03} & \textbf{9.71} \\
& ClimaX baseline     & 9.41           & 11.11          & 11.45          & 12.68          & 11.16 \\
& AirCast       & 10.72          & 12.67          & 12.21          & 13.40          & 12.25 \\
\addlinespace
\multirow{3}{*}{PM$_{10}$} 
& \textbf{TopoFlow (Ours)} & \textbf{14.78} & \textbf{17.06} & \textbf{17.70} & \textbf{19.45} & \textbf{17.25} \\
& ClimaX baseline     & 17.00          & 19.62          & 20.36          & 22.37          & 19.84 \\
& AirCast       & 17.90          & 20.29          & 19.85          & 21.63          & 19.92 \\
\addlinespace
\multirow{3}{*}{SO$_2$} 
& \textbf{TopoFlow (Ours)} & \textbf{1.44}  & \textbf{1.49}  & \textbf{1.48}  & \textbf{1.75}  & \textbf{1.54} \\
& ClimaX baseline     & 1.66           & 1.71           & 1.70           & 2.01           & 1.77 \\
& AirCast       & 2.89           & 2.88           & 2.81           & 3.27           & 2.96 \\
\addlinespace
\multirow{3}{*}{NO$_2$} 
& \textbf{TopoFlow (Ours)} & \textbf{7.79}  & \textbf{7.67}  & \textbf{7.99}  & \textbf{8.94}  & \textbf{8.10} \\
& ClimaX baseline     & 8.96           & 8.82           & 9.19           & 10.28          & 9.31 \\
& AirCast       & 9.75           & 9.16           & 9.33           & 10.13          & 9.59 \\
\addlinespace
\multirow{3}{*}{CO} 
& \textbf{TopoFlow (Ours)} & \textbf{72.11} & \textbf{72.50} & \textbf{74.97} & \textbf{81.81} & \textbf{75.35} \\
& ClimaX baseline     & 82.93          & 83.38          & 86.22          & 94.08          & 86.65 \\
& AirCast       & 100.34         & 100.89         & 104.33         & 113.84         & 104.85 \\
\addlinespace
\multirow{3}{*}{O$_3$} 
& \textbf{TopoFlow (Ours)} & \textbf{14.49} & \textbf{12.83} & \textbf{14.48} & \textbf{15.46} & \textbf{14.31} \\
& ClimaX baseline     & 16.66          & 14.75          & 16.65          & 17.78          & 16.46 \\
& AirCast       & 24.06          & 19.97          & 21.13          & 21.44          & 21.65 \\
\midrule
\multirow{3}{*}{\textbf{Overall}} 
& \textbf{TopoFlow (Ours)}  & \textbf{19.80} & \textbf{20.20} & \textbf{21.10} & \textbf{22.91} & \textbf{21.04} \\
& ClimaX baseline     & 22.77          & 23.23          & 24.26          & 26.53          & 24.20 \\
& AirCast       & 27.61          & 27.64          & 28.28          & 30.62          & 28.54 \\
\bottomrule
\end{tabular}
\end{table}

\subsection{Validation against independent ground stations}

\begin{table}[H]
\centering
\caption{\textbf{Validation Against Independent Ground Stations (2019).} RMSE in $\mu$g/m$^3$. \colorbox{green!15}{Green} = TopoFlow (Ours). \textbf{Bold} = best per lead time. $\downarrow$\% = relative increase vs. best model.}
\label{tab:stations_2019}
\footnotesize
\renewcommand{\arraystretch}{1.0}
\begin{tabular}{@{}llcccc|cc@{}}
\toprule
\textbf{Pollutant} & \textbf{Model} & \textbf{12h} & \textbf{24h} & \textbf{48h} & \textbf{96h} & \textbf{Average} & \textbf{$\downarrow$\%} \\
\midrule
\multirow{6}{*}{\textbf{PM$_{2.5}$}} 
& \cellcolor{green!15}\textbf{TopoFlow (Ours)} & \cellcolor{green!15}\textbf{26.9} & \cellcolor{green!15}\textbf{27.1} & \cellcolor{green!15}\textbf{28.0} & \cellcolor{green!15}\textbf{29.9} & \cellcolor{green!15}\textbf{28.0} & \cellcolor{green!15}-- \\
& ClimaX baseline        & 30.4 & 30.6 & 31.6 & 33.8 & 31.6 & +13\% \\
& AirCast       & 31.5 & 31.7 & 32.8 & 35.0 & 32.8 & +17\% \\
& Aurora      & 33.2 & 31.2 & 35.1 & 41.1 & 35.2 & +26\% \\
& CUACE$^\dagger$     & -- & 34--48 & -- & -- & 41.0 & +46\% \\
& CAMS$^\ddagger$   & 52.7 & 53.2 & 54.3 & 56.8 & 54.3 & +94\% \\
\addlinespace
\multirow{5}{*}{\textbf{PM$_{10}$}} 
& \cellcolor{green!15}\textbf{TopoFlow (Ours)} & \cellcolor{green!15}\textbf{49.4} & \cellcolor{green!15}\textbf{47.6} & \cellcolor{green!15}\textbf{47.2} & \cellcolor{green!15}\textbf{51.0} & \cellcolor{green!15}\textbf{48.8} & \cellcolor{green!15}-- \\
& ClimaX baseline        & 55.8 & 53.8 & 53.3 & 57.6 & 55.1 & +13\% \\
& AirCast       & 57.8 & 55.7 & 55.2 & 59.7 & 57.1 & +17\% \\
& Aurora      & 55.1 & 58.4 & 64.0 & 75.5 & 63.3 & +30\% \\
& CAMS$^\ddagger$   & 76.4 & 77.2 & 78.7 & 82.3 & 78.6 & +61\% \\
\addlinespace
\multirow{5}{*}{\textbf{SO$_2$}} 
& \cellcolor{green!15}\textbf{TopoFlow (Ours)} & \cellcolor{green!15}\textbf{13.1} & \cellcolor{green!15}\textbf{13.0} & \cellcolor{green!15}13.5 & \cellcolor{green!15}13.4 & \cellcolor{green!15}\textbf{13.3} & \cellcolor{green!15}-- \\
& ClimaX baseline        & 14.8 & 14.7 & 15.3 & 15.1 & 15.0 & +13\% \\
& AirCast       & 15.3 & 15.2 & 15.8 & 15.7 & 15.5 & +17\% \\
& Aurora      & 35.6 & 22.9 & \textbf{13.8} & \textbf{13.2} & 21.4 & +61\% \\
& CAMS$^\ddagger$   & 80.9 & 81.7 & 83.4 & 87.2 & 83.3 & +526\% \\
\addlinespace
\multirow{5}{*}{\textbf{NO$_2$}} 
& \cellcolor{green!15}\textbf{TopoFlow (Ours)} & \cellcolor{green!15}\textbf{21.9} & \cellcolor{green!15}24.2 & \cellcolor{green!15}24.2 & \cellcolor{green!15}\textbf{25.4} & \cellcolor{green!15}\textbf{23.9} & \cellcolor{green!15}-- \\
& Aurora      & 22.4 & \textbf{22.5} & \textbf{23.6} & 27.2 & 23.9 & 0\% \\
& ClimaX baseline        & 24.7 & 27.3 & 27.3 & 28.7 & 27.0 & +13\% \\
& AirCast       & 25.6 & 28.3 & 28.3 & 29.7 & 28.0 & +17\% \\
& CAMS$^\ddagger$   & 29.1 & 29.4 & 30.0 & 31.4 & 30.0 & +26\% \\
\addlinespace
\multirow{5}{*}{\textbf{CO}} 
& \textbf{Aurora} & \textbf{541} & \textbf{549} & 608 & 667 & \textbf{591} & -- \\
& \cellcolor{green!15}TopoFlow (Ours) & \cellcolor{green!15}585 & \cellcolor{green!15}598 & \cellcolor{green!15}\textbf{603} & \cellcolor{green!15}\textbf{616} & \cellcolor{green!15}601 & \cellcolor{green!15}+2\% \\
& CAMS$^\ddagger$   & 587 & 593 & 605 & 633 & 605 & +2\% \\
& ClimaX baseline        & 661 & 676 & 681 & 696 & 679 & +15\% \\
& AirCast       & 684 & 700 & 706 & 721 & 703 & +19\% \\
\addlinespace
\multirow{5}{*}{\textbf{O$_3$}} 
& \textbf{CUACE}$^\dagger$ & -- & \textbf{40--63} & -- & -- & \textbf{51.5} & -- \\
& Aurora      & \textbf{56.1} & 58.9 & \textbf{56.3} & \textbf{60.2} & 57.9 & +12\% \\
& CAMS$^\ddagger$   & 71.1 & 71.8 & 73.3 & 76.7 & 73.2 & +42\% \\
& \cellcolor{green!15}TopoFlow (Ours) & \cellcolor{green!15}79.4 & \cellcolor{green!15}75.7 & \cellcolor{green!15}75.8 & \cellcolor{green!15}76.0 & \cellcolor{green!15}76.7 & \cellcolor{green!15}+49\% \\
& ClimaX baseline        & 89.7 & 85.5 & 85.7 & 85.9 & 86.7 & +68\% \\
\midrule
\multirow{5}{*}{\textbf{Mean RMSE}} 
& \cellcolor{green!15}TopoFlow (Ours)  & \cellcolor{green!15}129.3 & \cellcolor{green!15}131.1 & \cellcolor{green!15}\textbf{132.1} & \cellcolor{green!15}\textbf{136.2} & \cellcolor{green!15}\textbf{132.2} & \cellcolor{green!15}-- \\
& Aurora      & \textbf{123.9} & \textbf{124.4} & 133.5 & 150.8 & 133.1 & +0.7\% \\
& ClimaX baseline        & 146.2 & 148.6 & 150.0 & 154.6 & 149.9 & +13\% \\
& CAMS$^\ddagger$   & 149.7 & 151.1 & 154.3 & 161.2 & 154.1 & +17\% \\
& AirCast       & 151.5 & 153.8 & 155.5 & 160.4 & 155.3 & +17\% \\
\bottomrule
\end{tabular}

\vspace{0.2cm}
\footnotesize
$^\dagger$ CUACE operational values from Dai et al. (2019) and Qi et al. (2022). \\
$^\ddagger$ CAMS forecast performance over China from Ge et al. (2020). \\
ClimaX baseline and AirCast were trained on CAQRA for fair comparison.
\end{table}

\begin{table}[H]
\centering
\caption{\textbf{Forecast errors relative to China's regulatory 
air quality thresholds (GB~3095-2012, Class~II, 24-hour 
average).} RMSE from independent OpenAQ validation (2019). 
Values in parentheses indicate RMSE as a percentage of the 
regulatory threshold. \colorbox{red!15}{Shaded cells} indicate errors exceeding 
50\% of the threshold, limiting operational utility for 
regulatory decision-making.}
\label{tab:threshold}
\small
\renewcommand{\arraystretch}{1.3}
\begin{tabular}{lcccccc}
\toprule
\textbf{Pollutant} & \textbf{Threshold} & \textbf{TopoFlow} & 
\textbf{ClimaX} & \textbf{AirCast} & \textbf{Aurora} & 
\textbf{CAMS} \\
\midrule
PM$_{2.5}$ & 75~$\mu$g/m$^3$ & 28.0 (37\%) & 31.6 (42\%) & 
32.8 (44\%) & 35.2 (47\%) & \cellcolor{red!15}54.3 (72\%) \\
PM$_{10}$ & 150~$\mu$g/m$^3$ & 48.8 (33\%) & 55.1 (37\%) & 
57.1 (38\%) & 63.3 (42\%) & \cellcolor{red!15}78.6 (52\%) \\
NO$_2$ & 80~$\mu$g/m$^3$ & 23.9 (30\%) & 27.0 (34\%) & 
28.0 (35\%) & 23.9 (30\%) & 30.0 (38\%) \\
SO$_2$ & 150~$\mu$g/m$^3$ & 13.3 (9\%) & 15.0 (10\%) & 
15.5 (10\%) & 21.4 (14\%) & \cellcolor{red!15}83.3 (56\%) \\
CO$^*$ & 4~mg/m$^3$ & 0.60 (15\%) & 0.68 (17\%) & 
0.70 (18\%) & 0.59 (15\%) & 0.61 (15\%) \\
O$_3$ & 160~$\mu$g/m$^3$ & 76.7 (48\%) & \cellcolor{red!15}86.7 (54\%) & 
-- & 57.9 (36\%) & 73.2 (46\%) \\
\bottomrule
\end{tabular}

\vspace{0.2cm}
\footnotesize
$^*$ CO RMSE reported in mg/m$^3$ to match the regulatory threshold unit (GB~3095-2012). Values in Supplementary Tables~5--6 are reported in $\mu$g/m$^3$ for consistency with other pollutants.
\end{table}

\subsection{Ablation study: tile granularity}

\begin{table}[H]
\centering
\caption{Tile granularity optimization for wind-guided reordering. Validation loss on 2017 holdout set.}
\label{tab:tile_ablation_supp}
\begin{tabular}{lccc}
\toprule
\textbf{Tiling Strategy} & \textbf{Wind Calculation} & \textbf{Val Loss} & \textbf{$\Delta$ Loss} \\
\midrule
Global wind direction & Single vector (country-wide) & 0.309 & --- \\
Coarse tiling & $2{\times}2$ tiles & 0.295 & $-$0.014 \\
 & $4{\times}4$ tiles & 0.293 & $-$0.016 \\
Medium tiling & $8{\times}8$ tiles & 0.291 & $-$0.018 \\
 & $12{\times}12$ tiles & 0.288 & $-$0.021 \\
 & $16{\times}16$ tiles & 0.280 & $-$0.029 \\
\rowcolor{green!10} \textbf{Fine tiling (optimal)} & $\mathbf{32{\times}32}$ \textbf{tiles} & \textbf{0.257} & \textbf{$-$0.052} \\
Over-parameterized & $16{\times}32$ tiles & 0.302 & +0.007 \\
 & $32{\times}64$ tiles & 0.376 & +0.067 \\
\bottomrule
\end{tabular}
\end{table}

The optimal tile size of $32 \times 32$ balances two competing considerations: (1) sufficient spatial extent to capture coherent wind patterns, and (2) adequate granularity to resolve mesoscale heterogeneity in the flow field. Smaller tiles lead to noisy wind estimates from insufficient averaging, while larger tiles smooth over important local variations.

\subsection{Ablation study: component contributions}

\begin{table}[H]
\centering
\caption{Component contribution analysis using optimal $32{\times}32$ tiles.}
\label{tab:component_ablation_supp}
\resizebox{\linewidth}{!}{
\begin{tabular}{lcccc}
\toprule
\textbf{Model} & \textbf{Scanning} & \textbf{Wind Tiles} & \textbf{Elevation $\alpha$} & \textbf{Val Loss} \\
\midrule
ClimaX baseline & Row-major & \texttimes & \texttimes & 0.264 \\
+ Wind-guided reordering & Wind-directed & $32{\times}32$ & \texttimes & 0.257 \\
\rowcolor{blue!10} \textbf{+ Topography-aware attention bias (TopoFlow)} & Wind-directed & $32{\times}32$ & \checkmark & \textbf{0.249} \\
\bottomrule
\end{tabular}}

\vspace{0.2cm}
\footnotesize
\textit{Notes:} Each row adds one component to the previous configuration. Wind-guided reordering reduces validation loss by 2.7\% relative to ClimaX baseline. Adding Topography-aware attention bias provides an additional 3.1\% reduction. The cumulative improvement is 5.7\% versus ClimaX baseline.
\end{table}

\section{Supplementary Discussion}

\subsection{Comparison with operational forecasting systems}

TopoFlow's performance advantage over operational systems (CAMS, CUACE) stems from fundamentally different modeling approaches. Numerical chemical transport models solve the advection-diffusion equation on three-dimensional grids, requiring explicit representation of emissions, chemistry, and deposition processes. These models are computationally expensive and sensitive to errors in emission inventories, which are particularly uncertain over China.

In contrast, TopoFlow learns the mapping from current atmospheric state to future pollutant concentrations directly from reanalysis data. This data-driven approach implicitly captures emission patterns, chemical transformations, and deposition processes without requiring explicit parameterization. The physics-guided attention mechanisms ensure that learned representations respect fundamental transport constraints.

\subsection{Fair comparison with AI baselines}

To ensure fair comparison, both ClimaX baseline and AirCast were trained on the same CAQRA dataset (2013--2016) with identical train/validation/test splits. This eliminates potential confounds from differences in training data quality, spatial coverage, or temporal extent. The performance improvements of TopoFlow over these baselines can therefore be attributed directly to the physics-guided architectural innovations rather than data advantages.

Aurora was not retrained on CAQRA as it is a foundation model with fixed pretrained weights. Its performance represents the zero-shot transfer capability of large-scale pretraining on global atmospheric data.

\subsection{Limitations of surface-only representation}

TopoFlow's limitation for O$_3$ and CO prediction reflects the importance of vertical atmospheric structure for these species. Surface ozone is influenced by stratospheric intrusions, where ozone-rich air descends through tropopause folding events. These processes are captured by Aurora's three-dimensional representation but cannot be resolved by TopoFlow's surface-only inputs.

Future extensions could incorporate pressure-level information from reanalysis products to capture vertical transport processes while maintaining the computational efficiency of the current architecture.

\subsection{Generalization to other regions}

While TopoFlow was trained and evaluated exclusively over China, the physics-guided mechanisms encode general principles that should transfer to other regions. Wind-guided attention aligns with advective transport structure regardless of geographic location. Topography-aware attention bias encodes universal mountain meteorology principles. The transformer architecture provides flexible capacity to adapt to different pollution regimes.

However, learned parameters (attention weights, embedding layers) may require fine-tuning for regions with substantially different emission patterns, meteorological regimes, or topographic configurations.


\end{document}